\documentclass{article}
\usepackage[numbers]{natbib}

\usepackage[preprint] {neurips_2024}

\usepackage[utf8]{inputenc} 
\usepackage[T1]{fontenc}    
\usepackage{hyperref}       
\usepackage{url}            
\usepackage{booktabs}       
\usepackage{amsfonts}       
\usepackage{nicefrac}       
\usepackage{microtype}      
\usepackage{xcolor}         

\usepackage{multirow} 
\usepackage{amsmath} %
\usepackage{xspace} %
\usepackage{titletoc} %
\usepackage{colortbl}

\usepackage{graphicx}
\usepackage{tcolorbox}
\tcbuselibrary{listings}
\usepackage{listings}

\usepackage{subcaption}

\newcommand{\huggingface}{\raisebox{-1.5pt}{\includegraphics[height=1.05em]{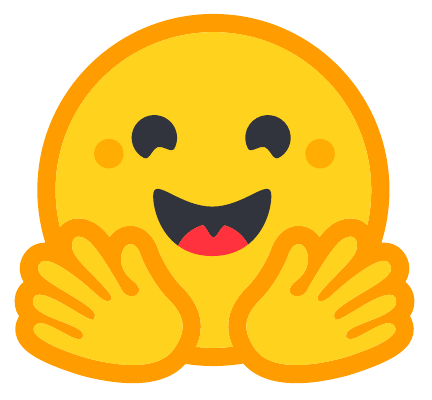}}\xspace}
\newcommand{\github}{\raisebox{-1.5pt}{\includegraphics[height=1.05em]{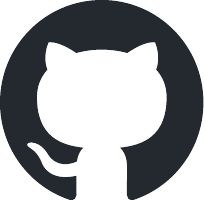}}\xspace}
\newcommand{\webpage}{\raisebox{-1.5pt}{\includegraphics[height=1.05em]{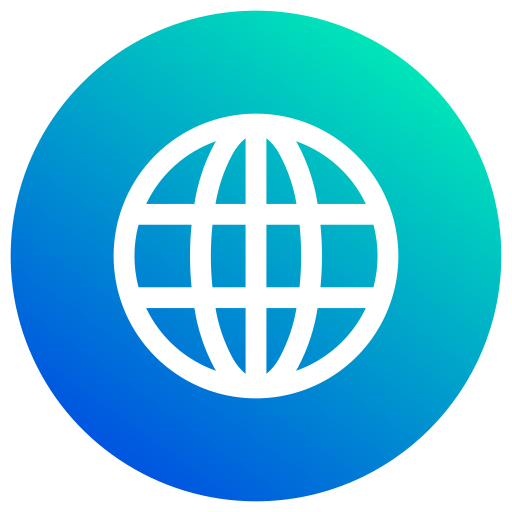}}\xspace}

\newcommand{\modelname}{LlamaV-o1\xspace}
\newcommand{\dataset}{VRC-Bench\xspace}

\title{
    \begin{minipage}{0.18\textwidth} 
        \centering
        \includegraphics[height=1.6cm]{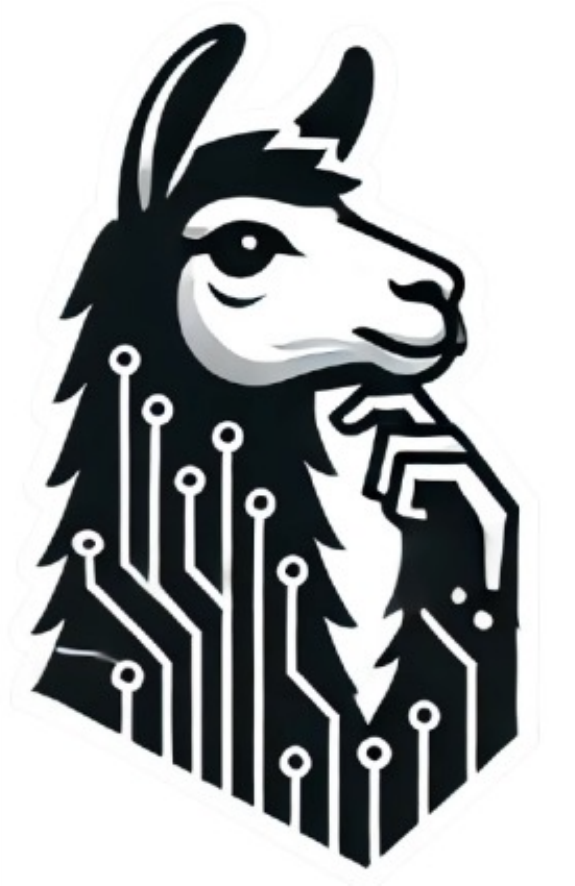} 
    \end{minipage}%
    \hspace{0.005cm} 
    \begin{minipage}{0.8\textwidth} 
        \raggedright 
        \textbf{LlamaV-o1: Rethinking Step-by-step \\ Visual Reasoning in LLMs}
    \end{minipage}
}

\author{
Omkar Thawakar\textsuperscript{\textnormal{1}}\thanks{Equal Contribution} 
\quad Dinura Dissanayake\textsuperscript{\textnormal{1*}} \quad Ketan More\textsuperscript{\textnormal{1*}}  
\quad Ritesh Thawkar\textsuperscript{\textnormal{1*}} \\
 \textbf{Ahmed Heakl}\textsuperscript{\textnormal{1*}} 
\quad \textbf{Noor Ahsan}\textsuperscript{\textnormal{1*}} 
\quad \textbf{Yuhao Li}\textsuperscript{\textnormal{1*}} 
\quad \textbf{Mohammed Zumri}\textsuperscript{\textnormal{1*}} 
\quad \textbf{Jean Lahoud}\textsuperscript{\textnormal{1*}} \\
 \textbf{Rao Muhammad Anwer}\textsuperscript{\textnormal{1}} 
\quad  \textbf{Hisham Cholakkal}\textsuperscript{\textnormal{1}} 
\quad  \textbf{Ivan Laptev}\textsuperscript{\textnormal{1}} 
\quad  \textbf{Mubarak Shah}\textsuperscript{\textnormal{2}} \\
 \textbf{Fahad Shahbaz Khan}\textsuperscript{\textnormal{1,3}} 
\quad  \textbf{Salman Khan}\textsuperscript{\textnormal{1,4}} \\
\\
\fontsize{10.5pt}{12pt}\selectfont
\textsuperscript{1}Mohamed bin Zayed University of AI, 
\textsuperscript{2}University of Central Florida, \\
\textsuperscript{3}Linköping University, 
\textsuperscript{4}Australian National University
}

\begin{document}

\maketitle

\begin{center}
\begin{tabular}{cll}
\webpage & \textbf{\modelname~ Project:} & {\small\url{https://mbzuai-oryx.github.io/LlamaV-o1/}} \\
\huggingface & \textbf{\modelname~ Model:} & {\small\url{https://huggingface.co/omkarthawakar/LlamaV-o1}} \\
\github & \textbf{\modelname~ Code:} & {\small\url{https://github.com/mbzuai-oryx/LlamaV-o1}}\\
\huggingface & \textbf{\dataset} & {\small\href{https://huggingface.co/datasets/omkarthawakar/VRC-Bench}{\nolinkurl{https://huggingface.co/datasets/omkarthawakar/VRC-Bench}}} 
\end{tabular}
\end{center}

\begin{abstract}
Reasoning is a fundamental capability for solving complex multi-step problems, particularly in visual contexts where sequential step-wise understanding is essential. Existing approaches lack a comprehensive framework for evaluating visual reasoning and do not emphasize step-wise problem-solving. 
To this end, we propose a comprehensive framework for advancing step-by-step visual reasoning in large language models (LMMs) through three key contributions.
\emph{First}, we introduce a visual reasoning chain benchmark specifically designed to evaluate multi-step reasoning tasks. The benchmark presents a diverse set of challenges with eight different categories ranging from complex visual perception to scientific reasoning with over 4k reasoning steps in total, enabling robust evaluation of LLMs’ abilities to perform accurate and interpretable visual reasoning across multiple steps. 
\emph{Second}, we propose a novel metric that assesses visual reasoning quality at the granularity of individual steps, emphasizing both correctness and logical coherence. The proposed metric offers deeper insights into reasoning performance compared to traditional end-task accuracy metrics.
\emph{Third}, we present a new multimodal visual reasoning model, named LlamaV-o1, trained using a multi-step curriculum learning approach, where tasks are progressively organized to facilitate incremental skill acquisition and problem-solving. The proposed LlamaV-o1 is designed for multi-step reasoning and learns step-by-step through a structured training paradigm. Extensive experiments show that our LlamaV-o1 outperforms existing open-source models and performs favourably against close-source proprietary models. Compared to the recent Llava-CoT, our LlamaV-o1 achieves an average score of 67.3 with an absolute gain of 3.8\% across six benchmarks while being 5$\times$ faster during inference scaling. Our benchmark, model, and code are publicly available. 
\end{abstract}

\section{Introduction}
Large Language Models (LLMs) are designed to understand and generate text, enabling them to handle a wide range of tasks such as translation~\cite{brown2020language,zhang2022opt,chowdhery2023palm,chen2025sharegpt4v,touvron2023llama,lu2024llamax}, summarization \cite{goyal2022news,zhang2024benchmarking,wang2023element}, and question-answering \cite{touvron2023llama,le2023bloom,touvron2023llama2}. 
The integration of visual data through Large Multimodal Models (LMMs) extends capabilities further by combining text, images, or videos, allowing for more complex multimodal tasks like image captioning, visual question answering, and video analysis.
To effectively solve these tasks, visual reasoning is essential for LMMs to process and connect diverse information, ensuring logical coherence and sequential problem-solving.
The ability to reason across multiple modalities is crucial to addressing complex real-world problems.

\begin{figure}[t!]
    \centering
    \includegraphics[width=\textwidth]{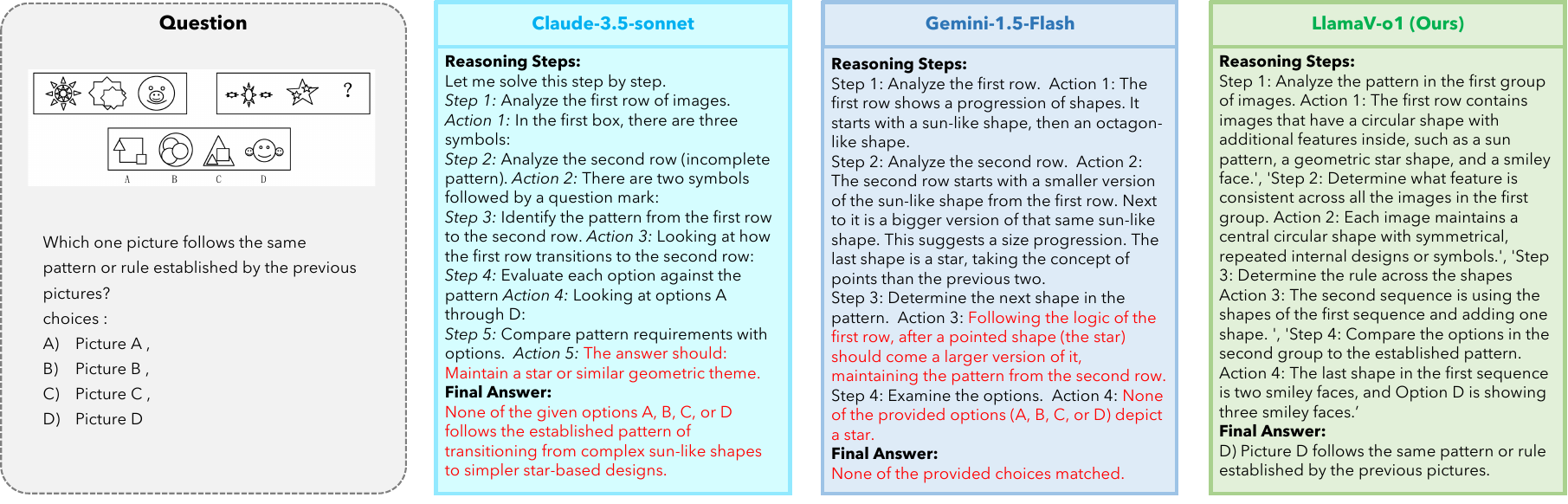} 
    \caption{
    Comparison of the reasoning abilities of our model (LlamaV-o1) with closed-source Gemini-1.5-Flash and Claude-3.5-Sonnet on an example in pattern recognition task from our proposed VRC-Bench. While Claude-3.5-Sonnet concludes "none of the options," its reasoning steps lack full alignment with the observed logic (highlighted in red). Gemini-1.5-Flash demonstrates weaker reasoning with less logical coherence details (highlighted in red). Our LlamaV-o1 provides better and more systematic reasoning, identifying that option D follows the established pattern, thereby showcasing its logical reasoning capability. Additional results are presented in Fig. \ref{fig:qual_comparison}.
    }
    \label{fig:teaser}
\end{figure}

To improve the problem-solving ability of LLMs, step-by-step reasoning is desired to break down complex tasks into easier components.
This approach resembles human cognitive processes, enabling models to track their thought processes and ensure logical consistency throughout their reasoning. 
By following a structured reasoning path, models can reach more accurate and interpretable conclusions.
To this end, previous works have demonstrated that prompting or fine-tuning LLMs to generate step-by-step rationales can lead to improvements in reasoning tasks \cite{lampinen2022tell,wei2022chain,wang2023selfconsistency}.
These methods encourage models to explicitly reason through each step, focusing on improving their ability to tackle complex tasks.
However, most existing works struggle to handle step-by-step multimodal reasoning tasks (see Fig. \ref{fig:teaser}).
Further, a notable gap in current visual reasoning benchmarks is their lack of emphasis on step-by-step reasoning.
Most benchmarks focus primarily on end-task accuracy, neglecting the quality of intermediate reasoning steps.
Moreover, the absence of standardized evaluation can likely lead to inaccurate comparisons between models, making it difficult to assess their true visual reasoning capabilities.

Our work strives to bridge the aforementioned gaps by introducing a holistic approach to evaluating step-by-step visual reasoning capabilities. 
To this end, we introduce Visual Reasoning-Chain (VRC-Bench), a comprehensive a benchmark specifically designed to assess multi-step visual reasoning tasks.
The benchmark spans 8 diverse categories: Visual Reasoning, Math \& Logic Reasoning, Social \& Cultural Context, Medical Imaging (Basic Medical Science), Charts \& Diagram Understanding, OCR \& Document Understanding, Complex Visual Perception, and Scientific Reasoning.
It includes over 1,000 challenging samples, meticulously curated to evaluate reasoning capabilities across various domains. Furthermore, the benchmark features 4,173 manually verified reasoning steps, ensuring accuracy and reliability in assessing step-by-step logical reasoning.
Recognizing that measuring end-task accuracy is insufficient, we present a new metric that evaluates visual reasoning quality at the granularity of individual steps, focusing on both correctness and logical coherence.
Further, we explore the combined advantages of integrating Beam Search with Multi-Step curriculum learning in the training of visual reasoning models. By leveraging the efficiency of Beam Search alongside the progressive structure of curriculum learning, the proposed model incrementally acquires skills, starting with simpler tasks such as summary of the approach and question derived captioning and advancing to more complex multi-step reasoning scenarios, ensuring both optimized inference and robust reasoning capabilities.
We observe this structured training paradigm not only to enhance the model's performance but also to improve interpretability and adaptability in handling different visual reasoning tasks (see Fig. \ref{fig:teaser}).
Our extensive experiments demonstrate that our approach, named LlamaV-o1, outperforms existing open-source methods, including the recent Llava-CoT \cite{llava_cot} model, across multiple evaluation metrics. 

In summary, this paper presents a unified framework that aims to advance step-by-step visual reasoning capabilities through a new benchmark, a novel metric, and a new model trained with curriculum learning. 
Our contributions are as follows:
\begin{itemize}
    \item Step-by-Step Visual Reasoning Benchmark: To the best of our knowledge, the proposed benchmark is the first effort designed to evaluate multimodal multi-step reasoning tasks across diverse topics. The proposed benchmark, named VRC-Bench, spans around eight different categories (Visual Reasoning, Math \& Logic Reasoning, Social \& Cultural Context, Medical Imaging (Basic Medical Science), Charts \& Diagram Understanding, OCR \& Document Understanding, Complex Visual Perception and Scientific Reasoning) with over 1,000 challenging samples and more than 4k reasoning steps. 
    \item Novel Evaluation Metric: A metric that assesses the reasoning quality at the level of individual steps, emphasizing both correctness and logical coherence.
    \item Combined Multi-Step Curriculum Learning and Beam Search Approach: A multimodal reasoning method, named LlamaV-o1, that combines the structured progression of curriculum learning with the efficiency of Beam Search. The proposed approach ensures incremental skill development while optimizing reasoning paths, enabling the model to be effective in complex multi-step visual reasoning tasks in terms of both accuracy and efficiency. Specifically, the proposed LlamaV-o1 achieves an absolute gain of 3.8\% in terms of average score across six benchmarks while being 5$\times$ faster, compared to the recent Llava-CoT~\cite{llava_cot}. 
\end{itemize}

\section{Related Works}

\textbf{Reasoning with LLMs:} The development of robust reasoning capabilities in Large Language Models (LLMs) has been a focal point of research.
Early work often relied on neural-symbolic methods for explicit modeling of the reasoning process using formal language instead of natural language \cite{roy-roth-2015-solving,chiang-chen-2019-semantically,amini-etal-2019-mathqa}. 
However, the emergence of powerful LLMs has prompted new approaches that leverage their inherent reasoning abilities \cite{wu2024comparative}.
For example, inference time computing is scaled in recent models to perform reasoning before giving the final answer \cite{xiong2024llava, weng2022large, huang2022large, openai2024o1}. 
Techniques like Chain-of-Thought (CoT) prompting, where a complex question is decomposed into intermediate reasoning steps, have shown promise in guiding LLMs to structured solutions \cite{wei2022chain,yao2024tree}.
Nevertheless, maintaining logical consistency, especially in tasks requiring multi-step inference, poses a significant challenge, leading to errors and hallucinated outputs \cite{xu2024llava,madaan2024self}. 
LLMs, even with CoT guidance, might generate unfaithful explanations, deviate from logical reasoning paths, and struggle with verifying and selecting correct reasoning steps \cite{wei2022chain}. These approaches are further extended to VLMs.

\textbf{Reasoning with VLMs:} 
Visual reasoning tasks require models to possess visual perception and high-level cognitive abilities \cite{gupta2023visual, khan2014geometry, xu2024llava}.
The visual reasoning skills have broad applicability across domains such as science \cite{lu2022learn}, mathematics \cite{lu2023mathvista}, robotic planning \cite{hu2023look} and advanced question answering \cite{yue2023mmmu}. 
Similar to the case of LLMs, the conventional approaches employed neural-symbolic methods to explicitly model the reasoning process \cite{garcez2019neural,vedantam2019probabilistic,andreas2016neural}. 
For example, \citep{amizadeh2020neuro} propose differentiable logic formalism to decouple the reasoning aspect of VQA from visual perception.
More recent VLMs leverage the reasoning capabilities of LLMs for visual tasks. Visual programming \cite{gupta2023visual} provides a modular neuro-symbolic system based on computer vision models as functions and GPT-3 LLM for compositional visual reasoning. Zhang et al. \cite{zhang2024improve} argue that VLM training with concise answers results in reduced generalization to more complex problems requiring reasoning. They use GPT-4o model to create rationales and use correct and incorrect reasoning chains in training to enhance model's reasoning ability via reinforcement learning (RL) \cite{rafailov2024direct}. In contrast, LlaVA-o1 \cite{xu2024llava} does not use RL and advocates for stage-wise reasoning instead of CoT prompting, where the answer is worked out sequentially via summarization, interpretation, reasoning, and conclusion steps.
Our work builds on \cite{xu2024llava} and shows the importance of curriculum learning and path search for visual reasoning.

\textbf{Benchmarks for Visual Reasoning:}
Several datasets and benchmarks have been developed to evaluate and advance visual reasoning in VLMs. These datasets vary in complexity, visual context, and reasoning skills required. Some notable examples are as follows. 
CLEVR (Compositional Language and Elementary Visual Reasoning) tests visual reasoning abilities like counting, comparisons, and logical inference through rendered images and automatically generated questions \cite{johnson2017clevr}. 
StrategyQA is a multi-hop question-answering dataset on Wikipedia that necessitates implicit decompositions and diverse reasoning strategies \cite{geva2021did}. 
ScienceQA offers a large-scale multimodal science dataset with multi-modal contexts, diverse science topics, and annotated answers with corresponding lectures and explanations \cite{lu2022learn}. 
A consolidated mathematical reasoning benchmark in diverse visual contexts called MathVista incorporates 28 existing multimodal datasets and three new datasets \cite{lu2023mathvista}. 
Zhang \emph{et al.} \cite{zhang2024improve} propose ShareGPT-4o-Reasoning, a comprehensive CoT dataset containing 193k examples covering various VQA tasks, designed to improve CoT reasoning in VLMs.
However, these benchmarks do not provide step-by-step reasoning in complex evaluation scenarios and generally judge the correctness based on only the final answer. In this work, our goal is to provide a comprehensive benchmark that assesses the reasoning chains as well as the final outcome in complex reasoning scenarios.

\begin{figure}[t]
    \centering
    \includegraphics[width=\textwidth]{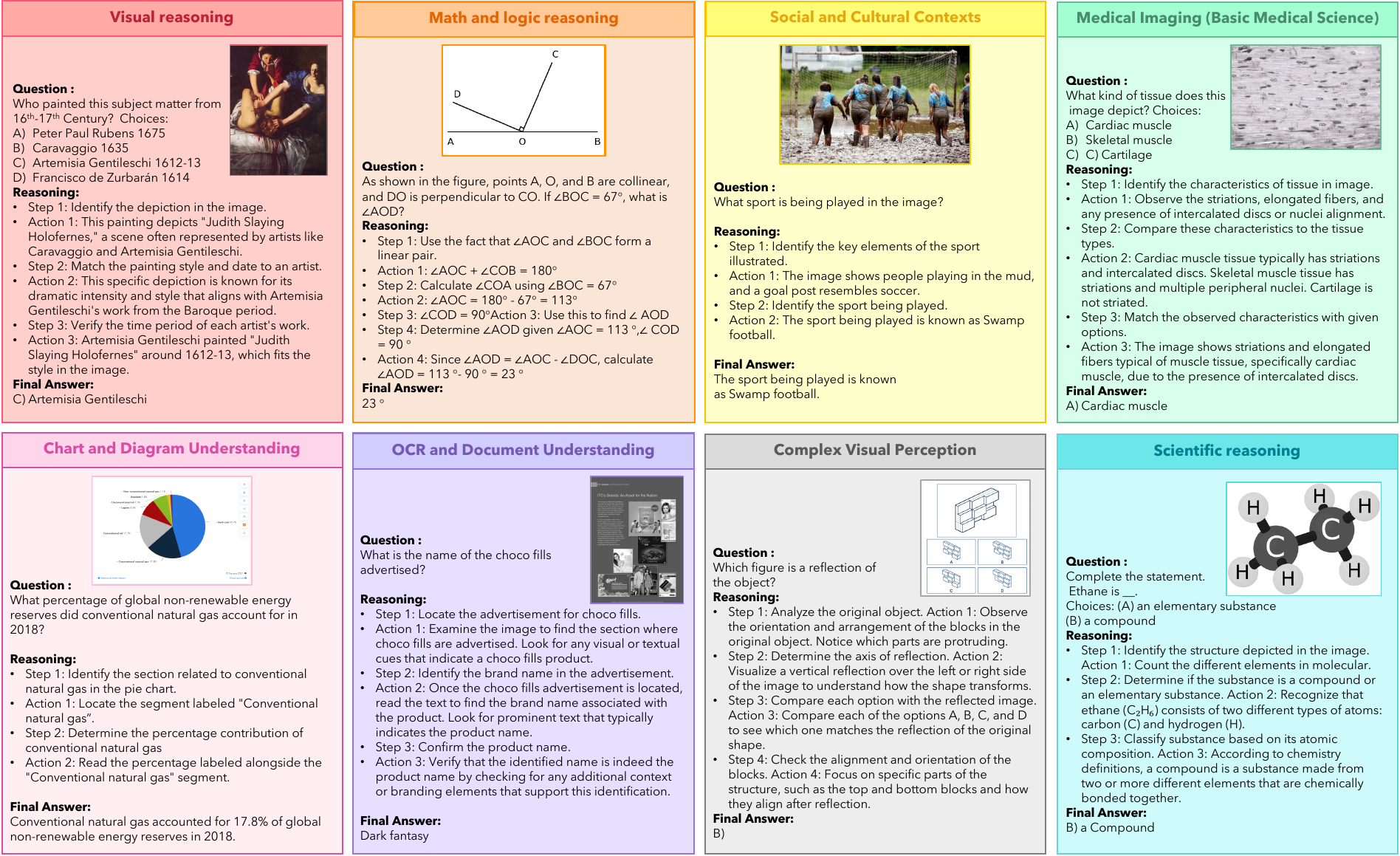} 
    \caption{
    The proposed VRC-Bench examples show the diverse and challenging reasoning tasks our benchmark encompasses, spanning a wide range of modalities and contexts. Each example emphasizes step-by-step reasoning, starting from task comprehension and progressing to logical inference and answer generation. The tasks include mathematical reasoning using geometric principles, scientific classification based on molecular structures, visual interpretation of charts and diagrams, artistic identification from historical paintings, and medical diagnosis from tissue images.
    For instance, one example demonstrates the calculation of an angle in a geometric diagram by leveraging linear pairs and perpendicular relationships. Another highlights scientific reasoning by identifying ethane as a compound based on its molecular composition. Visual perception tasks challenge the model to analyze pie charts for global energy reserves or recognize reflected shapes. Artistic and cultural tasks require identifying paintings and sports based on visual and contextual cues. Finally, tasks in medical imaging and advertisement recognition test the model's ability to classify tissue types or extract product names through careful observation.
    }
    \label{fig:dataset_examples}
\end{figure}

\section{Step-by-Step Visual Reasoning Benchmark: VRC-Bench}
To facilitate a thorough assessment of the reasoning capabilities in complex scenarios, we introduce a step-by-step visual reasoning benchmark.
This benchmark serves as a structured tool to assess both the logical progression of reasoning chains and the accuracy of the final outcomes generated by LMMs.
By integrating diverse datasets that include a diverse range of topics, such as science, mathematics, medical knowledge, social sciences, and data interpretation, we ensure that our evaluation benchmark captures the diverse aspects of reasoning.

\begin{figure}[t]
    \centering
    \includegraphics[width=0.95\textwidth]{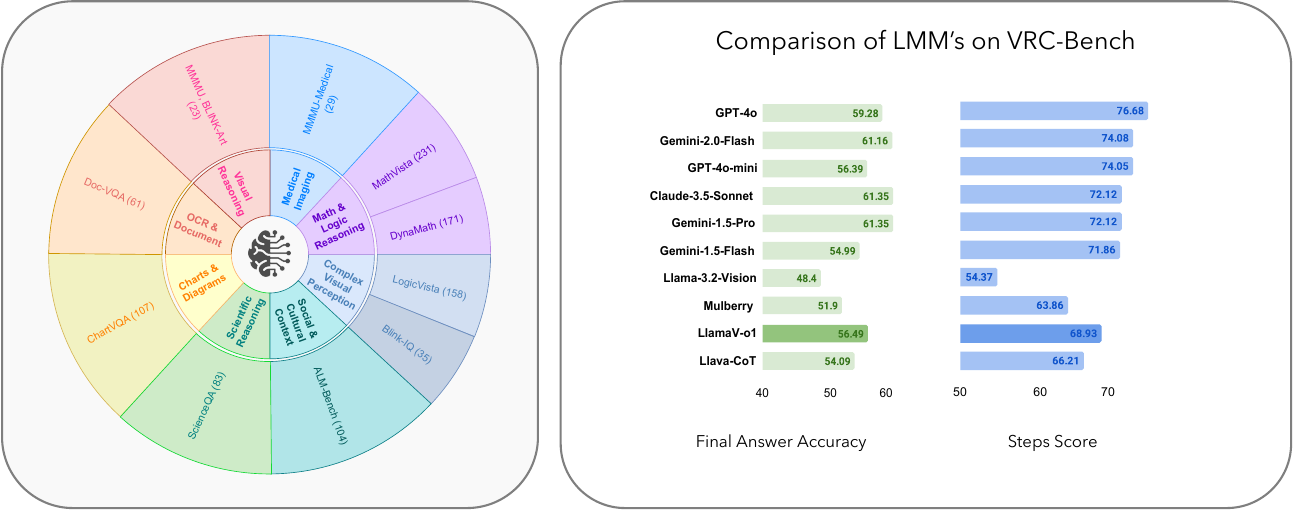} 
    \caption{
    The figure illustrates our comprehensive benchmark structure and comparative performance of LMMs on the proposed ReasoningChain-Bench. \emph{(Left)} The dataset spans multiple domains, including carefully selected samples for mathematical and logical reasoning (e.g., MathVista \cite{lu2024mathvista} with 231 samples and LogicVista with 158 samples), scientific reasoning (e.g., Science-QA \cite{lu2022learn} with 83 samples), and visual perception (e.g., Blink-IQ-Test \cite{fu2024blink} with 35 samples). Additionally, it includes specialized areas such as medical imaging (e.g., MMMU-Medical \cite{yue2023mmmu} with 29 samples), cultural and social understanding (e.g., ALM-Bench \cite{vayani2024alm} with 104 samples), and document understanding through OCR (e.g., Doc-VQA \cite{mathew2020docvqa} with 61 samples). By integrating tasks like chart and diagram comprehension (e.g., Chart-VQA \cite{masry2022chartqa} with 107 samples), our dataset not only covers a broad spectrum of real-world applications but also expand LMM's ability to reason, perceive, and interpret complex multimodal information. 
    \emph{(Right)} The bar chart compares various SoTA reasoning  models on the VRC-Bench, highlighting both final answer accuracy and step-by-step reasoning scores. The models evaluated for complex reasoning tasks include GPT-4o, Gemini-2.0-Flash, Claude-3.5-Sonnet, and Llava-CoT. Our benchmark evaluates models not only on their ability to generate accurate final answers but also on the coherence and logical flow of their reasoning steps. Our approach, LlamaV-o1, outperforms GPT-4o-mini, Gemini-1.5-Flash and Llava-CoT in the VRC-Bench, achieving superior results in final answer accuracy across complex multimodal reasoning tasks.
    }
    \label{fig:data_split_chart}
\end{figure}

\subsection{Benchmark Creation}
\textbf{Benchmark Domains:} To ensure a comprehensive assessment of reasoning capabilities, our step-by-step visual reasoning benchmark incorporates samples from several specific datasets across various domains.
Figure \ref{fig:dataset_examples} shows examples of the questions and answers included in our benchmark.
The data distribution is shown in Figure \ref{fig:data_split_chart}.
By integrating these diverse sources, we capture a wide range of reasoning scenarios, allowing for an extensive evaluation of the models' abilities to respond to complex inquiries.
Based on these diverse data samples, we generate step-by-step reasoning steps using a semi-automated annotation pipeline with detailed rationales. Next, we outline the main domains covered in the benchmark and then explain the annotation process.   

\textit{Mathematical and Logical Reasoning}:
This category includes datasets focused on mathematical and logical tasks. 
MathVista \cite{lu2024mathvista}, provides a variety of mathematical problems, while DynaMath \cite{zou2024dynamic} offers dynamic mathematical challenges.
Additionally, ChartQA \cite{masry2022chartqa} encompasses tasks related to chart and diagram comprehension, allowing evaluation of visual reasoning in logical contexts.

\textit{Scientific Reasoning}:
For scientific reasoning, we include samples from Science-QA \cite{lu2022learn} to test the model's ability to answer questions based on scientific knowledge and reasoning. 
Furthermore, MMMU-Medical \cite{yue2023mmmu}, focuses on medical imaging tasks, assessing the model's capability in interpreting complex multimodal medical data.

\textit{Cultural and Social Understanding}:
To assess the model’s ability to recognize and interpret diverse cultural scenarios, we include samples from ALM-Bench \cite{vayani2024alm}, which is designed to assess understanding of the social and cultural context.

\textit{Other Visual Reasoning Scenarios}:
We further include samples from other visual reasoning datasets.
LogicVista \cite{xiao2024logicvistamultimodalllmlogical} and Blink-IQ \cite{fu2024blink} focus on complex visual perception, providing challenges that require the model to analyze and interpret intricate visual information. 
Doc-VQA \cite{mathew2020docvqa} targets OCR and document understanding, evaluating the model's ability to extract information from text-based documents. 
Lastly, MMMU \cite{yue2023mmmu} and BLINK \cite{fu2024blink} (Art Split) contribute to visual reasoning tasks.

\textbf{Semi-Automatic Step-by-Step Reasoning Generation:} We adopt a semi-automatic approach to generate step-by-step reasoning responses. 
We begin by using the GPT-4o model to create detailed reasoning steps and answers for the various questions in our dataset. 
This involves crafting specific prompts to guide the model in producing detailed logical reasoning. 
In this way, we efficiently generate various reasoning chains with a consistent format, where the step-by-step reasoning includes all the required steps and actions to reach the desired answer. Additional details are provided in Appendix (Section.~\ref{appendix:generating_reasoning_steps_from_target_models}).

\textbf{Manual Verification:} Since automated responses are not always reliable, we perform manual verification to ensure that all reasoning steps are accurate and correct.
In this stage, a team of verifiers meticulously reviewed the generated reasoning chains and final answers, making necessary adjustments to enhance clarity and correctness. Our benchmark consists of examples spanning over 8 diverse categories as shown in Fig.~\ref{fig:data_split_chart}. 
We ask the verifiers to add missing reasoning steps when necessary, and we drop examples with less than three reasoning steps after the verification except some samples from MathVista as they can be addressed with 2 steps. Over 25\% of the data was corrected during the manual verification resulting in more than 1,000 samples and carefully verified 4,173 reasoning steps.  
The manual verification stage is essential for establishing a trustworthy ground truth, which serves as the benchmark for evaluating LMMs performance in our evaluations.

\begin{table}[ht]
\centering
\caption{
An overview of comprehensive set of attributes considered in our evaluation to assess the quality of reasoning in LMMs. These attributes focus on critical aspects such as faithfulness, informativeness, and logical coherence of reasoning steps. Key measures include ensuring alignment of reasoning steps with the source (Faithfulness-Step and Token), completeness of information (Informativeness-Step), and identifying issues like hallucinations, redundancy, or missing steps. Additional metrics, such as Semantic Coverage and Reasoning Alignment, evaluate the logical and semantic integrity of the response. Together, these metrics provide a robust framework for evaluating the accuracy, completeness, and reliability of LLM-generated reasoning.
}
\label{tab:evaluation_metrics}
\begin{tabular}{@{}p{0.3\textwidth}p{0.65\textwidth}@{}}
\toprule
\textbf{Metric}                & \textbf{Definition}                                                                                          \\
\midrule
\texttt{Faithfulness-Step}              & Measures how well the reasoning steps in the LMM response align with the source reasoning steps.             \\
\texttt{Faithfulness-Token}             & Extends Faithfulness-Step to token-level granularity, checking if the content within each step is accurate.   \\
\texttt{Informativeness-Step}          & Measures how well the reasoning steps extract all relevant information from the context.                      \\
\texttt{Repetition-Token}               & Identifies repeated or unnecessarily paraphrased reasoning steps.                                            \\
\texttt{Hallucination}                 & Detects irrelevant or fabricated reasoning steps not aligned with the source.                                \\
\texttt{Redundancy}                     & Identifies redundant reasoning steps that do not add value.                                                  \\
\texttt{Semantic Coverage-Step}         & Measures how well the response covers the essential semantic elements of the source reasoning steps.         \\
\texttt{Reasoning Alignment}            & Overall alignment between the hypothesis and reference reasoning chain.                                      \\
\texttt{Commonsense}                    & Checks for missing commonsense reasoning are required to solve the problem.                                      \\
\texttt{Missing Step}                   & Identifies if any necessary reasoning steps are missing.                                                     \\
\bottomrule
\end{tabular}
\end{table}

\subsection{Evaluation Framework}
\label{eval_benchmark}
While multiple previous methods have been proposed for evaluating reasoning chains \cite{golovnevaroscoe,prasad2023receval}, these methods exhibit various limitations. 
These methods adopt a reference-free approach, as they do not depend on a set ground truth.
While this allows for flexibility in evaluation, it can lead to significant problems.
For example, even if the reasoning steps are logically sequenced, a minor error can lead to a major disruption in the reasoning chain but still result in a high score.
This compromises the accuracy of the assessment, as it does not truly reflect the quality of the reasoning.
In our research, we stress on the importance of having a ground truth for scoring.
By comparing generated responses to a reliable reference, we aim to improve the accuracy of our evaluations.
Using a ground truth reasoning chain allows us to better identify and address inaccuracies.

\textbf{Evaluation Metric:} To overcome the shortcomings of reference-free metrics, we use GPT-4o~\cite {openai2024gpt4ocard} to compare the predictions generated by the model against a ground truth.
This method allows us to evaluate reasoning quality using specific metrics that focus on different aspects of alignment and accuracy.
We base our metric on the reference-free ROSCOE suite of metrics \cite{golovnevaroscoe} and propose a reference-based metric.
We show the details of the measures used in our metric in Table \ref{tab:evaluation_metrics}.
For instance, we use the Faithfulness-Step and Faithfulness-Token metrics to assess how well the reasoning aligns with the source.
The Faithfulness-Step metric scores alignment on a scale from 1 to 10, providing clear feedback on the accuracy of each reasoning step.
We also measure Informativeness-Step, which checks if all critical information is included.
By incorporating attributes like Hallucination and Redundancy, we can spot irrelevant or repetitive reasoning that reduces clarity.
The final scoring process averages all attribute scores to give a comprehensive evaluation of reasoning quality. Additional details such as system prompt used for scoring are provided in Appendix (Section.~\ref{appendix:system_prompt_for_reasoning_eval}).

\section{Proposed Step-by-Step Visual Reasoning Model: LlamaV-o1}

Our proposed approach introduces several key contributions to advance multimodal reasoning in LMMs. First, we leverage curriculum learning to train the model progressively, starting with foundational tasks like summarization of approach and question based caption generation before advancing to detailed, multi-step reasoning. This structured approach helps the model manage complexity, improve logical coherence, and generalize effectively to challenging scenarios. Second, we scale inference efficiently with a simple yet effective Beam Search technique, which generates multiple beams in parallel and selects the most optimal one, ensuring both efficiency and high-quality outputs. This method significantly reduces computational costs, achieving constant scaling compared to the linear scaling of traditional approaches in terms of model calls.

\subsection{Curriculum Learning for Large Multimodal Models}
LMMs are powerful tools for understanding and generating content across different data types, such as text, images, and video. However, reasoning in such models, especially in complex multi-step scenarios, presents unique challenges. Models often struggle to handle step-by-step reasoning because reasoning requires not only understanding the input but also maintaining consistency and logical clarity across multiple steps. This is where curriculum learning becomes an essential strategy.
Curriculum learning, inspired by human education systems, involves training a model progressively, starting from simpler tasks and gradually introducing more complex ones. This approach has shown significant benefits in improving model performance across various tasks, particularly when the tasks require reasoning over multiple modalities. For instance, curriculum learning has been successfully applied in multimodal learning such as Visual Question Answering (VQA)~\cite{kembhavi2017you} and captioning tasks~\cite{johnson2016densecap}. These studies demonstrate that models trained with simpler examples first and later gradually increasing task difficulty can generalize better to more complex problems.

Curriculum learning is a powerful approach for enhancing reasoning capabilities in LMMs by adopting a progressive training strategy. Starting with complex task and gradually introducing more difficult complex challenges, it helps models to build foundational skills incrementally. In the case of multimodal models, this structured progression allows to manage complexity effectively, as they first learn to interpret basic relationships between modalities, such as connecting text with images, before tackling more intricate scenarios. By ensuring a strong basis for logical reasoning, curriculum learning improves the coherence of multi-step tasks, enabling models to maintain consistency and alignment across steps. In addition, curriculum learning addresses challenges like catastrophic forgetting, which can occur when models are directly fine-tuned on complex tasks, leading to overfitting and poor generalization. By focusing initially on simpler tasks, models consolidate fundamental patterns before progressing to more advanced problems. This approach mirrors human learning, where basic skills are mastered before tackling complex concepts, ensuring better generalization and adaptability. Overall, curriculum learning establishes a robust framework for developing reasoning capabilities, making multimodal models more reliable and effective across a wide range of real-world applications.

\subsection{Multi-Step Chain-of-Thought for Improved Reasoning}
Multi-step chain-of-thought reasoning is crucial for tackling complex tasks that require sequential decision-making and logical coherence. Unlike single-step reasoning, which often overlooks intermediate steps, multi-step reasoning allows models to break down problems into smaller, manageable parts, ensuring transparency and consistency throughout the process. This step-by-step approach mirrors how humans solve complex problems by systematically reasoning through each step. For instance, answering a multifaceted question about an image might involve identifying objects, understanding their relationships, and synthesizing this information to form a coherent answer. Embracing multi-step reasoning not only enhances the interpretability of multimodal models but also aligns them closer to human-like problem-solving, setting the stage for more robust and versatile AI systems.

\subsubsection{Multi-Step Reasoning Stages}
Multi-step reasoning in chain-of-thought (CoT) frameworks enables models to break down complex tasks into incremental steps, mirroring the human approach to problem-solving. This structured process ensures that each reasoning step is explicit and logical, improving transparency and accuracy. Our multi-step reasoning comprises of following steps also depicted as an example in the box below:

\begin{itemize}
    \item \textbf{Task Understanding:} The model begins by understanding the question and the context.
    \item \textbf{Task Summarization:} The next step involves generating a summary of the visual data to ensure the model has a holistic understanding. This stage prepares the model to focus on relevant action items to be taken to get the final answer.
    \item \textbf{Detailed Caption Generation:} To narrow the scope further, the model generates a detailed caption, which identifies specific labels and their corresponding values in the chart. This step ensures that the model accurately interprets the visual elements.
    \item \textbf{Logical Reasoning:} The model then formulates a logical reasoning process to locate and interpret the required data. This reasoning step breaks the task into sub-goals, ensuring a systematic approach. 
    \item \textbf{Final Answer Generation:} Finally, the model outputs the final answer based on the reasoning process and the extracted context.
\end{itemize}

Breaking down reasoning into these stages ensures that models process complex queries methodically, reducing errors and enhancing interpretability. multi-step training reinforces the ability to handle each stage effectively, enabling LMMs to perform better on tasks that require step-by-step logical reasoning. This approach not only improves accuracy but also provides a transparent pathway for verifying and refining the model's outputs.

\begin{tcolorbox}[colback=white,colframe=black!45,
    fonttitle=\bfseries, title=Multi-Step Training Example ,
    listing options={basicstyle=\ttfamily\small,breaklines=true},
    ] 
\small

\begin{center}\includegraphics[width=0.7\linewidth]{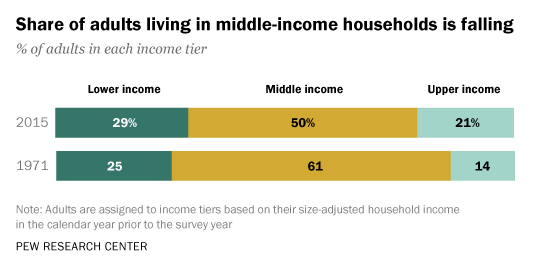}\end{center}

\textbf{Human:} \texttt{What was the value of the middle-income share in 1971? Answer the question using a single word or phrase. Please generate a summary of the picture.} 

\textbf{Assistant:} I will examine the image to find the relevant data for the middle-income share in 1971 and present the answer in the specified format.

\textbf{Human:} \texttt{Please generate a detailed caption for the image.}

\textbf{Assistant:} The image displays a bar chart comparing the percentage of adults in the lower, middle, and upper-income tiers for the years 2015 and 1971. It shows that in 1971, the middle-income share was 61\%. 

\textbf{Human:} \texttt{Please generate a detailed reasoning to answer the question given the caption.}

\textbf{Assistant:}  To solve the problem, I will look for the specific percentage associated with the middle-income group in the year 1971 from the provided chart. The image clearly shows that in 1971, the middle-income tier accounted for 61\% of adults. 

\textbf{Human:} \texttt{Please generate the final answer based on reasoning steps. Do not output anything else.}

\textbf{Assistant:} 0.61 

\textbf{Human:} \texttt{How many did the share of the middle-class change from 1971 to 2015? Answer the question using a single word or phrase. Please generate a summary of the picture.}

\textbf{Assistant:} To solve this problem, I will analyze the image data to determine the change in the share of adults living in middle-income households from 1971 to 2015. I will compare the percentages provided for each year and calculate the difference.

\textbf{Human:} \texttt{Please generate a detailed caption for the image.}

\textbf{Assistant:} The image is a bar graph from Pew Research Center showing the share of adults in lower, middle, and upper-income households for the years 1971 and 2015. For 1971, 61\% were in the middle income, and for 2015, 50\% were in the middle income.

\textbf{Human:} \texttt{Please generate a detailed reasoning to answer the question given the caption.}

\textbf{Assistant:} To find the change in the share of middle-class adults from 1971 to 2015, subtract the 2015 percentage from the 1971 percentage. The middle-income share in 1971 was 61\%, and in 2015 it was 50\%. Thus, the change is calculated as 61\% - 50\% = 11\%. The difference in share, expressed as a decimal, is 0.11. 

\textbf{Human:} \texttt{Please generate the final answer based on reasoning steps. Do not output anything else.}

\textbf{Assistant:} 0.11

\end{tcolorbox}

\subsubsection{Data Preparation and Model Training}
To implement our curriculum learning strategy effectively, we divide the model training process into two stages, each designed to incrementally enhance the model's reasoning capabilities while ensuring a robust understanding of multimodal inputs. This structured approach allows the model to acquire foundational reasoning skills in the first stage and progressively refine its ability to provide detailed, step-by-step answers in the second stage.

\noindent \textbf{Stage 1: Training for Summarization and Caption Generation:} In the first stage, the model is trained to generate two critical components: (1) a summary of the approach needed to answer the question and (2) a detailed caption describing the relevant aspects of the input data, such as visual elements in an image. The training data for this stage is derived from 18K samples from the Cap-QA split of PixMo dataset \cite{deitke2024molmo} and 57K samples from Geo170K dataset from G-LLaVa~\cite{gao2023g}. Each sample includes a question paired with input data (e.g., an image or chart). Cap-QA split of PixMo dataset contains examples having grounded captions based on input question whereas the Geo170K dataset contains question-answer pairs with their reasoning steps. This stage ensures that the model learns to contextualize the input and outline a high-level reasoning plan before diving into detailed steps.
The focus in this stage is to help the model gather the structure of reasoning tasks, improving its ability to decouple the problem into simpler elements. By focusing on structured training, the model develops the ability to handle multi-step tasks while maintaining a clear and organized flow of thought.

\noindent \textbf{Stage 2: Training for Detailed Reasoning and Final Answer Generation:} In the second stage, the model builds upon the foundation established in Stage 1. Here, the model is trained not only to generate the summary and caption but also to provide detailed reasoning based on these components. Finally, the model outputs the correct answer derived from the reasoning process. For this stage, we use the original Llava-CoT dataset \cite{llava_cot}, which contains 99K structured samples comprising of various domains such as General VQA and Science-Targeted VQA from multiple sources. The General VQA data sources includes ShareGPT4V~\cite{chen2025sharegpt4v}, ChartQA~\cite{masry2022chartqa}, A-OKVQA~\cite{schwenk2022okvqa}, DocVQA~\cite{mathew2021docvqa}, PISC~\cite{junnan2017people}, CLEVR~\cite{johnson2017clevr} whereas, Science-Targeted VQA sources consists GeoQA+~\cite{cao2022augmented}, AI2D~\cite{kembhavi2016diagram}, ScienceQA~\cite{lu2022learn} and CLEVR-Math~\cite{lindstrom2022clevr} respectively.   Each sample consists of a summary, caption, detailed reasoning, and the final answer.
The training process in this stage involves multi-step interactions, where the model progressively learns to break down the approach into incremental reasoning steps. This incremental learning ensures that the model refines its logical flow and systematically integrates the information from summaries and captions into actionable reasoning steps.

In second stage, the multi-step training methodology is key to the model's success. During Stage 1, the model learns to organize its thoughts and outline a strategy, effectively setting the stage for the detailed reasoning required in Stage 2. By the time the model reaches Stage 2, it is already equipped with the ability to outline a structured approach, making it easier to focus on breaking down complex tasks into step-by-step solutions. This approach improves the interpretability, accuracy, and robustness of the model, enabling it to excel in complex multimodal reasoning tasks. Our results demonstrate that by leveraging datasets such as PixMo and Llava-CoT in a curriculum learning framework, the model can effectively transition from high-level problem understanding to detailed, step-by-step reasoning, achieving state-of-the-art performance in multi-step reasoning benchmarks.

\noindent \textbf{Model Training:} We leverage the PixMo and LLaVA-CoT-100k datasets to train our model using a curriculum learning strategy combined with a Supervised Fine-Tuning (SFT) approach. For this work, we select Llama-3.2-11B-Vision-Instruct~\cite{llama3.2} as our base model due to its strong foundation in multimodal reasoning and instruction-following capabilities. The fine-tuning process involves full-parameter optimization, allowing the model to adapt effectively to the structured reasoning tasks provided by the PixMo and LLaVA-CoT-100k dataset.
Training is performed on a high-performance computing node equipped with 8 NVIDIA A100 (80GB) GPUs, ensuring efficient handling of the large-scale dataset and the computational requirements of the model. During the initial stage of curriculum learning, the model is fine-tuned on the PixMo dataset to develop foundational reasoning skills, such as generating summaries and captions.
Further training details, such as the number of epochs, learning rate, optimizer settings, and batch size, are outlined in Appendix for reproducibility of our work.

\subsubsection{Optimizing Inference Efficiency: Beam Search}
Inference efficiency is a critical factor in deploying large multimodal models for real-world applications, particularly when handling complex reasoning tasks. To address this, we adopt a Beam Search strategy that significantly improves inference efficiency and reasoning quality compared to existing approaches like LLava-CoT~\cite{xu2024llava}. Our method is designed to balance computational complexity with output quality, enabling faster and more reliable inference.

\noindent \textbf{Simplified Output Design:} Unlike LLava-CoT~\cite{xu2024llava}, our approach does not require a highly structured output format. This flexibility simplifies the reasoning process, allowing the model to focus on generating high-quality outputs without the overhead of rigid structural constraints. This design choice makes our method more adaptable to a wide range of reasoning scenarios, improving generalization across tasks.

\noindent \textbf{Improved Efficiency with Beam Search:} The Beam Search technique allows us to generate multiple reasoning paths in parallel and select the most optimal one. This approach enhances both the quality and consistency of the model's outputs. By evaluating multiple candidates and selecting the best, we ensure that the final answer is logical and robust. One of the key advantages of our method is its computational efficiency. The inference time scaling of our approach has a complexity of $O(n)$, which is significantly more efficient than LLava-CoT's $O(n^2)$ scaling. This linear complexity ensures that our method is scalable to larger datasets and more complex reasoning tasks without a proportional increase in computational cost.

\begin{table}[t]
\centering
\caption{Comparison of models based on Final Answer accuracy and Reasoning Steps performance on the proposed VRC-Bench. The best results in each case (closed-source and open-source) are in bold. Our LlamaV-o1 achieves superior performance compared to its open-source counterpart (Llava-CoT) while also being competitive against the closed-source models.
}
\label{tab:model_comparison}
\resizebox{\textwidth}{!}{
\begin{tabular}{l|cccccc|cccc}
\toprule
& \multicolumn{6}{c}{\textit{Close-Source}} & \multicolumn{4}{c}{\textit{Open-Source}} \\
\textbf{Model} & \textbf{GPT-4o} & \textbf{Claude-3.5} & \textbf{Gemini-2.0} & \textbf{Gemini-1.5} & \textbf{Gemini-1.5} & \textbf{GPT-4o} & \textbf{Llama-3.2} & \textbf{Mulberry} & \textbf{Llava-CoT} & \textbf{ LlamaV-o1} \\
& ~\cite{openai2024gpt4ocard} & \textbf{Sonnet}~\cite{claude3.5sonnet_blog} & \textbf{Flash} & \textbf{Pro}~\cite{gemini1.5} & \textbf{Flash}~\cite{gemini1.5} & \textbf{mini}~\cite{gpt4omini} & \textbf{Vision}~\cite{llama3.2} & ~\cite{yao2024mulberry} & ~\cite{llava_cot} & (Ours) \\
\midrule
\textbf{Final Answer} & 59.28 & \textbf{61.35} & 61.16 & \textbf{61.35} & 54.99 & 56.39 & 48.40 & 51.90 & 54.09 & \textbf{56.49} \\
\textbf{Steps} & \textbf{76.68} & 72.12 & 74.08 & 72.12 & 71.86 & 74.05 & 58.37 & 63.86 & 66.21 & \textbf{68.93} \\
\bottomrule
\end{tabular}
}
\end{table}

\begin{figure}[t]
    \centering
    \includegraphics[width=\textwidth]{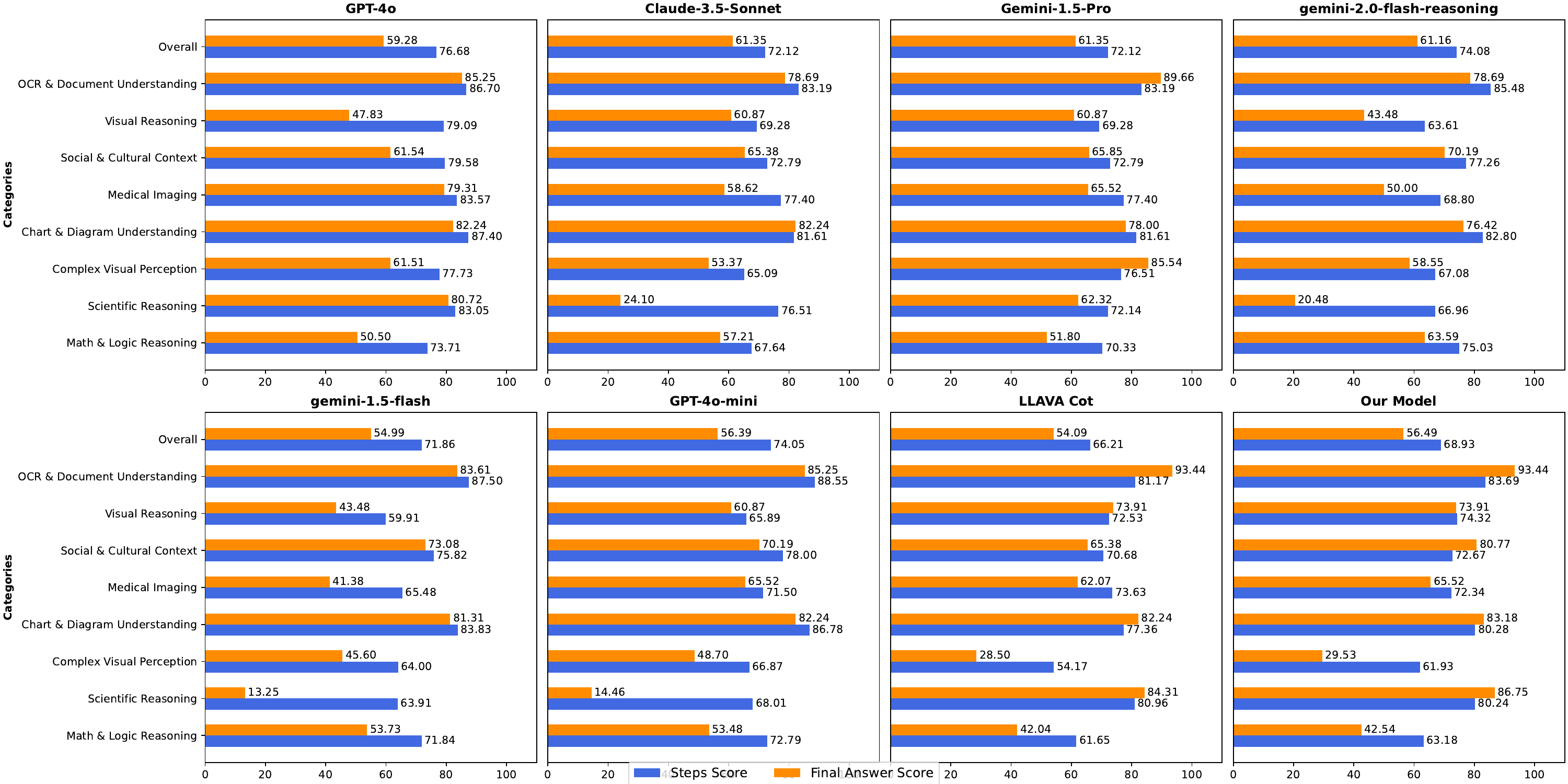} 
    \caption{
    The comprehensive comparison of category-wise and overall performance scores achieved by various models on diverse reasoning tasks. The evaluation spans multiple domains, including Math \& Logic Reasoning, Scientific Reasoning, Complex Visual Perception, Chart \& Diagram Understanding, Medical Imaging, Social \& Cultural Context, Visual Reasoning, and OCR \& Document Understanding. The models assessed include GPT-4o, Claude-3.5-Sonnet, Gemini variants, LLAVA-CoT, and our proposed model.
    Our model demonstrates consistently superior performance in critical categories such as Math \& Logic Reasoning, Chart \& Diagram Understanding, and Medical Imaging, achieving a balanced improvement across both step-by-step reasoning (Step Scores) and final answer accuracy (Final Answer Scores). Compared to LLAVA-CoT, our approach excels in maintaining high accuracy across tasks while showcasing robustness and interpretability in multi-step reasoning challenges.
    }
    \label{fig:category_comparison}
\end{figure}

\section{Experiments}

In this section, we evaluate the performance of our proposed model trained with a curriculum learning strategy. We use Llama-3.2-11B-Vision-Instruct~\cite{llama3.2} as the baseline model, leveraging its robust vision-language understanding as a foundation. Training is conducted using the PixMo~\cite{deitke2024molmo} subset and LLaVA-CoT-100k~\cite{llava_cot} datasets, carefully curated to support step-by-step reasoning tasks. The curriculum learning framework ensures the model progresses from simpler tasks, such as generating summaries and captions, to more complex multi-step reasoning tasks involving detailed logic and final answer generation.

To assess the effectiveness of our approach, we evaluate the model on our newly proposed reasoning benchmark, designed to test multi-step chain-of-thought capabilities in multimodal contexts. Additionally, we benchmark our model against six commonly used multimodal benchmarks employed in LLaVA-CoT~\cite{llava_cot}, covering diverse domains such as visual reasoning, mathematical problem-solving, and scientific reasoning. This comprehensive evaluation demonstrates the robustness of our method, highlighting significant improvements in both reasoning quality and inference efficiency compared to the baseline.

\subsection{Experimental Setup}
For our experiments, we use Llama-3.2-11B-Vision-Instruct as the baseline model, fine-tuned using the llama-recipes framework in a Supervised Fine-Tuning (SFT) manner. This robust foundation enables us to implement our curriculum learning strategy effectively, training the model on reasoning tasks that progressively increase in complexity. The training datasets include PixMo and LLaVA-CoT-100k, which we specifically tailored to support multi-step reasoning tasks and detailed chain-of-thought explanations.

For evaluation, we utilize a diverse set of benchmarks, including both our proposed reasoning benchmark and six established multimodal benchmarks: MMStar~\cite{mmstar}, MMBench~\cite{mmbench}, MMVet~\cite{mmvet}, MathVista~\cite{mathvista}, AI2D~\cite{ai2d}, and Hallusion~\cite{hallusionbench}. These benchmarks comprehensively evaluate the model’s capabilities across general visual question answering, mathematical and scientific reasoning, and handling language hallucinations and visual illusions. Our proposed benchmark is designed to assess step-by-step reasoning and final answer as described in the Section~\ref{eval_benchmark}, and evaluations are conducted using a fuzzy evaluation strategy where GPT-4o acts as the judge, ensuring robust assessments of generated outputs. To maintain consistency and fairness, we adopt the VLMEvalKit~\cite{vlmevalkit} framework, as used in LLaVA-CoT, to evaluate all models on six established multimodal benchmarks. This open-source toolkit ensures reproducibility and allows for direct comparison of performance metrics across different models. Our rigorous experimental setup highlights the effectiveness of our approach in advancing multimodal reasoning capabilities.

\subsection{Results and Discussion}
Our model demonstrates significant improvements over existing methods on our proposed reasoning benchmark, as shown in Table \ref{tab:model_comparison}. The evaluation compares final answer accuracy and step-by-step reasoning performance across state-of-the-art models. While models like GPT-4o, Claude-3.5-Sonnet and Gemini-2.0-Flash exhibit strong reasoning capabilities, our approach achieves the better final answer accuracy (56.49) compared to GPT-4o-mini and LLava-CoT and competitive step scores (68.93\%). This highlights the model’s ability to generate accurate outputs while maintaining logical coherence in multi-step tasks. The structured curriculum learning strategy and effective inference design have been critical in achieving these results.

Figure \ref{fig:category_comparison} illustrates the category-wise performance of our model compared to leading reasoning models in various domains from our benchmark such as Math \& Logic, Scientific Reasoning, and Complex Visual Perception. Our model outperforms others in several challenging categories, including Chart \& Diagram Understanding (83.18\%), Scientific Reasoning (86.75\%) and OCR \& Document Understanding (93.44\%). These improvements outline the model’s ability to handle tasks requiring visual and logical reasoning in accordance. The results also highlight balanced performance across all categories, reflecting the versatility of our approach.

Table \ref{tab:model_performance} summarizes the performance of various models on six established benchmarks: MMStar, MMBench, MMVet, MathVista, AI2D, and HallusionBench. Among open-source models, our method achieves the highest average score of 67.33\%, surpassing other prominent models like LLaVA-CoT (63.50). Notably, our model demonstrates significant strengths in reasoning-intensive benchmarks such as MMVet (65.40\%) and Hallusion (63.51\%). These results validate the effectiveness of our model in handling diverse and complex multimodal tasks.

\begin{figure}[t!]
    \centering
    \includegraphics[width=0.95\textwidth]{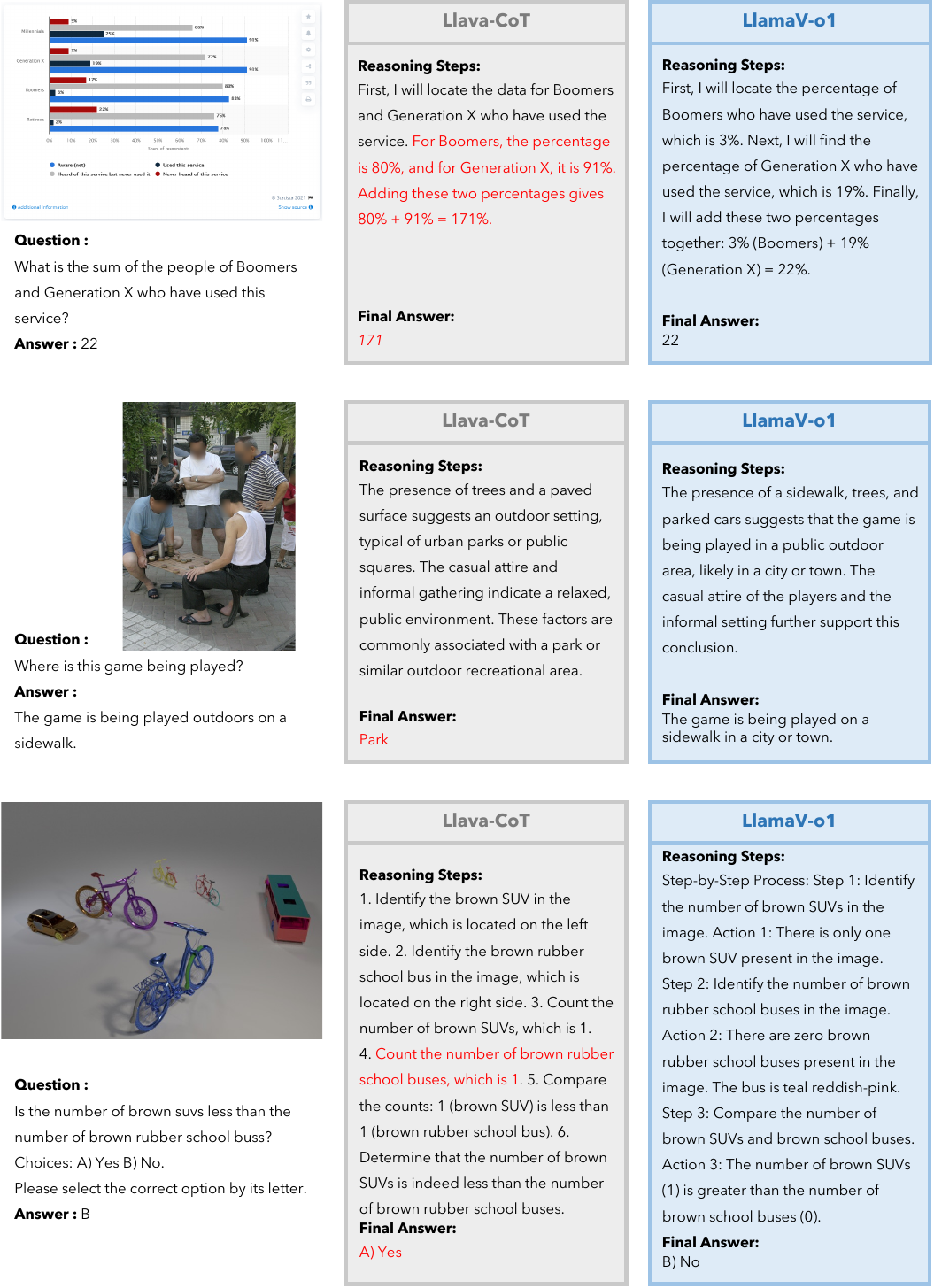} 
    \caption{
    Qualitative comparison between Llava-CoT and the proposed LlamaV-o1 on examples from the VRC-Bench. \emph{First row:} the example shows visual reasoning capabilities on an example chart. Here, Llava-CoT makes mistakes (highlighted in red) for both the intermediate steps and the final answer. In Comparison, our LlamaV-o1 provides an accurate description of the steps as well as the final answer. \emph{Second row:} While both Llava-CoT and our LlamaV-o1 provide accurate step descriptions on an example real-world VQA, Llava-CoT fails to infer the final answer. \emph{Last row:} Llava-CoT fails to accurately answer for the counting task, while also missing the intermediate counting steps. In contrast, our LlamaV-o1 model performs better in intermediate reasoning steps while also providing the accurate final answer. 
    }
    \label{fig:qual_comparison}
\end{figure}

The results demonstrate that our approach outperforms recent open-source visual reasoning methods while achieving favorable results against its close-source counterparts. By leveraging curriculum learning and optimizing inference efficiency with Beam Search, our model effectively balances reasoning accuracy and computational complexity. Our performance improvements in reasoning tasks are complemented by robust handling of logical errors and visual illusions, as evidenced in benchmarks like HallusionBench. Fig.~\ref{fig:qual_comparison} presents a qualitative comparison between the recent Llava-CoT and our LlamaV-o1 on different examples from the VRC-Bench. Our LlamaV-o1 achieves superior performance both in reasoning steps and the final answer, compared to Llava-CoT.

\begin{table}[t]
\centering
\caption{Performance comparison on six benchmark datasets (MMStar~\cite{mmstar}, MMBench~\cite{mmbench}, MMVet~\cite{mmvet}, MathVista~\cite{mathvista}, AI2D~\cite{ai2d}, and Hallusion~\cite{hallusionbench}) along with their average scores. The comparison includes both close-source and open-source models. The best performing close-source model is GPT-4o with an average score of 71.8\%. Among open-source models, our proposed LlamaV-o1 achieves the best performance with an average score of 67.33\% outperforming the recent Llava-CoT by 3.8\%. 
}
\label{tab:model_performance}
\small
\setlength{\tabcolsep}{7pt}
    \resizebox{\columnwidth}{!}{%
    \begin{tabular}{lccccccc}
    \toprule
    \textbf{Model} & \textbf{MMStar} & \textbf{MMBench} & \textbf{MMVet} & \textbf{MathVista} & \textbf{AI2D} & \textbf{Hallusion} & \textbf{Average} \\
    \midrule
    \multicolumn{7}{l}{\textbf{Close-Source}} \\
    GPT-4o-0806~\cite{openai2024gpt4ocard} & 66.0 & 82.4 & 80.8 & 62.7 & 84.7 & 54.2 & 71.8 \\
    Claude3.5-Sonnet-0620~\cite{claude3.5sonnet_blog} & 64.2 & 75.4 & 68.7 & 61.6 & 80.2 & 49.9 & 66.7 \\
    Gemini-1.5-Pro~\cite{gemini1.5} & 56.4 & 71.5 & 71.3 & 57.7 & 79.1 & 45.6 & 63.6 \\
    GPT-4o-mini-0718~\cite{gpt4omini} & 54.9 & 76.9 & 74.6 & 52.4 & 77.8 & 46.1 & 63.8 \\
    \midrule
    \multicolumn{7}{l}{\textbf{Open-Source}} \\
    InternVL2-8B~\cite{chen2024internvl} & 62.50 & 77.40 & 56.90 & 58.30 & 83.60 & 45.00 & 64.00 \\
    Ovis1.5-Gemma2-9B~\cite{lu2024ovis} & 58.70 & 76.30 & 50.90 & 65.60 & 84.50 & 48.20 & 64.00 \\
    MiniCPM-V2.6-8B~\cite{yao2024minicpm} & 57.10 & 75.70 & 56.30 & 60.60 & 82.10 & 48.10 & 63.30 \\
    Llama-3.2-90B-Vision-Inst~\cite{llama3.2} & 51.10 & 76.80 & 74.10 & 58.30 & 69.50 & 44.10 & 62.30 \\
    VILA-1.5-40B~\cite{liu2024nvila} & 53.20 & 75.30 & 44.40 & 49.50 & 77.80 & 40.90 & 56.90 \\
    Mulberry-7B~\cite{yao2024mulberry} & 61.30 & 75.34 & 43.90 & 57.49 & 78.95 & 54.10 & 62.78 \\
    Llava-CoT~\cite{llava_cot} & 57.60 & 75.00 & 60.30 & 54.80 & 85.70 & 47.80 & 63.50 \\
    \midrule
    \multicolumn{7}{l}{\textbf{Our Models}} \\
    Llama-3.2-11B-Vision-Inst~\cite{llama3.2} (baseline) & 49.80 & 65.80 & 57.60 & 48.60 & 77.30 & 40.30 & 56.90 \\
    \rowcolor{orange!25}
    \textbf{LlamaV-o1 (Ours)} & 59.53 & 79.89 & 65.40 & 54.40 & 81.24 & 63.51 & 67.33 \\
    \bottomrule
    \end{tabular}
    }
\end{table}

\begin{table}[t]
\centering
\caption{
Impact of our proposed contributions on multimodal reasoning tasks across six benchmarks: MMStar, MMBench, MMVet, MathVista, AI2D, and Hallusion. Starting with Curriculum Learning combined with Multi-Step CoT reasoning (2nd row), the model achieves a 9.14\% absolute gain compared to base model Llama-3.2-11B-Vision-Inst~\cite{llama3.2}, demonstrating its ability to handle complex multi-step reasoning effectively. This baseline approach leverages structured training to improve performance across diverse tasks, including logical reasoning and visual understanding. By incorporating Beam Search, the model's performance further improves (3rd row). This enhancement is particularly noticeable in benchmarks such as MMVet (65.40\% vs. 61.88\%), MathVista (54.40\% vs. 53.20\%), and AI2D (81.24\% vs. 80.18\%), showcasing the model's ability to generalize better with more accurate reasoning. 
Our final approach that combines curriculum learning with optimized inference strategies achieves an overall average gain of 10.43\%, compared to the baseline.
}
\label{tab:contribution_of _components}
\small
\setlength{\tabcolsep}{4pt}
    \resizebox{\columnwidth}{!}{%
    \begin{tabular}{lccccccc}
    \toprule
    \textbf{Model} & \textbf{MMStar} & \textbf{MMBench} & \textbf{MMVet} & \textbf{MathVista} & \textbf{AI2D} & \textbf{Hallusion} & \textbf{Average} \\
    \midrule
    Llama-3.2-11B-Vision-Inst (\textit{baseline}) & 49.80 & 65.80 & 57.60 & 48.60 & 77.30 & 40.30 & 56.90 \\
    \rowcolor{orange!10}
    + Curriculum with Multi-Step CoT Reasoning & 58.13 & 79.55 & 61.88 & 53.20 & 80.18 & 63.31 & 66.04 \\
    \rowcolor{orange!15}
    + Beam Search & 59.53 & 79.89 & 65.40 & 54.40 & 81.24 & 63.51 & 67.33 \\
    \bottomrule
    \end{tabular}
    }
\end{table}

\subsection{Ablations}
\noindent \textbf{Impact of Proposed Components:} The Table~\ref{tab:contribution_of _components} showcases the impact of our proposed components of LlamaV-o1 on improving performance in complex visual reasoning tasks across six multimodal benchmarks: MMStar, MMBench, MMVet, MathVista, AI2D, and Hallusion. Starting with a curriculum learning strategy combined with multi-step Chain-of-Thought (CoT) reasoning, the model achieves an average score of 66.08\%, demonstrating its ability to handle reasoning-intensive tasks effectively. By incorporating Beam Search, which optimizes the selection of reasoning paths, the performance further improves, achieving the highest average score of 67.33\%. This improvement is particularly significant in benchmarks such as MMVet (65.40\% vs. 61.88\%), MMStar (59.53\% vs. 58.13\%), and AI2D (81.24\% vs. 80.18\%), which evaluate the model's logical, visual, and contextual reasoning abilities. These results highlight the effectiveness of combining progressive training with optimized inference, enabling the model to generalize better across complex tasks and consistently deliver accurate and coherent reasoning. This validates the importance of our proposed contributions in advancing multimodal reasoning systems.

\begin{table}[ht]
    \centering
    \caption{
    Comparison of inference scaling techniques on the MMVet benchmark, evaluated using a single NVIDIA A100 GPU.
    \textbf{Left:} Llava-CoT with stage-level beam search shows improved MMVet scores with more beams but suffers from quadratic scaling, significantly increasing inference time.
    \textbf{Right:} {Performance of our approach utilizing Beam Search} achieving higher MMVet scores with much lower inference time, due to its linear scaling efficiency.
    For instance, our method scores 65.40 with four beams in 6.1 GPU hours, compared to Llava-CoT’s 62.9 score requiring 46.1 GPU hours. This demonstrates the efficiency and practicality of our approach for real-world applications.
    }
    \label{tab:ablation}
    \begin{minipage}{0.49\textwidth}
        \setlength{\tabcolsep}{3pt}
        \resizebox{\textwidth}{!}{%
        \begin{tabular}{l|ccc}
        \toprule
            \textbf{Inference Scaling} & \textbf{\# Beams} & \textbf{MMVet Score} & \textbf{Time (GPU Hours)} \\
            \toprule
            No Scaling & 1 & 60.3 & 3.8 \\
            Stage-level  & 2 & 61.7 & 20.1 \\
            Stage-level  & 3 & 62.3 & 38.5 \\
            Stage-level  & 4 & 62.9 & 54.1 \\
            \bottomrule
        \end{tabular}
        }
    \end{minipage}
    \hfill
    \begin{minipage}{0.49\textwidth}
    \setlength{\tabcolsep}{3pt}
        \resizebox{\textwidth}{!}{%
        \begin{tabular}{l|cccc}
        \toprule
            \textbf{Inference Scaling} & \textbf{\# Beams} & \textbf{MMVet Score} & \textbf{Time (GPU Hours)} \\
            \toprule
            No Scaling & 1 & 63.63 & 2.7 \\
            Beam Search & 2 & 64.26 & 4.8 \\
            Beam Search & 3 & 64.92 & 5.7 \\
            Beam Search & 4 & 65.40 & 6.1 \\
            \bottomrule
        \end{tabular}
        }
    \end{minipage}
\end{table}

\noindent \textbf{Effectiveness of Inference Scaling Techniques:} The Table~\ref{tab:ablation} focuses on the efficiency and effectiveness of inference scaling techniques on the MMVet benchmark. We compare the newly introduced stage-level beam search used in Llava-CoT with Beam Search in our proposed approach. Both approaches are evaluated based on MMVet scores and inference time, measured on a single NVIDIA A100 GPU (80GB).
\textbf{Stage-Level Beam Search (Llava-CoT):} Increasing the number of beams improves the MMVet score incrementally (from 60.3\% with 1 beam to 62.9\% with 4 beams). However, this improvement comes at a significantly higher computational cost due to linear scaling (time complexity of $O(n)$) based on model calls, with inference time rising from 3.8 GPU hours for 1 beam to 54.1 GPU hours for 4 beams. This scaling inefficiency limits the practicality of the stage-level approach for real-world applications.
\textbf{Beam Search (Ours):} In contrast, our method achieves significantly better MMVet scores while maintaining a constant scaling (time complexity of $O(1)$) of inference time in terms of model calls. With 1 beam, our model already outperforms Llava-CoT (63.63\% vs. 60.3\%). As the number of beams increases, the MMVet score improves further, reaching 65.40 with 4 beams in just 4.2 GPU hours, a fraction of the computational cost of Llava-CoT. This demonstrates that Beam Search is not only more efficient with higher accuracy but also suitable for real-world applications.

\section{Conclusion}
In this paper, we proposed a comprehensive approach for advancing multimodal reasoning by introducing a new benchmark, a novel metric, and an innovative model trained using curriculum learning. Our model demonstrates significant improvements over existing methods, achieving state-of-the-art performance on challenging benchmarks while maintaining efficiency in inference. The incorporation of curriculum learning enabled the model to develop foundational reasoning skills progressively, resulting in improved generalization and robustness across diverse tasks. Our results highlight the effectiveness of our design choices, including the structured training strategy, efficient inference mechanism, and rigorous evaluation using both our benchmark and widely recognized datasets. Our contributions aim to provide a capable multimodal reasoning model, emphasizing the importance of interpretable, step-by-step reasoning in improving AI's ability to handle complex, multi-step tasks.

\bibliographystyle{plain}
\bibliography{neurips_2024}

\begin{thebibliography}{10}

\bibitem{claude3.5sonnet_blog}
Claude 3.5 sonnet, 2024.
\newblock Available at: \url{https://www.anthropic.com/news/claude-3-5-sonnet}.

\bibitem{openai2024gpt4ocard}
OpenAI (2024).
\newblock Gpt-4o system card, 2024.

\bibitem{amini-etal-2019-mathqa}
Aida Amini, Saadia Gabriel, Shanchuan Lin, Rik Koncel-Kedziorski, Yejin Choi, and Hannaneh Hajishirzi.
\newblock {M}ath{QA}: Towards interpretable math word problem solving with operation-based formalisms.
\newblock In Jill Burstein, Christy Doran, and Thamar Solorio, editors, {\em Proceedings of the 2019 Conference of the North {A}merican Chapter of the Association for Computational Linguistics: Human Language Technologies, Volume 1 (Long and Short Papers)}, pages 2357--2367, Minneapolis, Minnesota, June 2019. Association for Computational Linguistics.

\bibitem{amizadeh2020neuro}
Saeed Amizadeh, Hamid Palangi, Alex Polozov, Yichen Huang, and Kazuhito Koishida.
\newblock Neuro-symbolic visual reasoning: Disentangling.
\newblock In {\em International Conference on Machine Learning}, pages 279--290. Pmlr, 2020.

\bibitem{andreas2016neural}
Jacob Andreas, Marcus Rohrbach, Trevor Darrell, and Dan Klein.
\newblock Neural module networks.
\newblock In {\em Proceedings of the IEEE conference on computer vision and pattern recognition}, pages 39--48, 2016.

\bibitem{brown2020language}
Tom Brown, Benjamin Mann, Nick Ryder, Melanie Subbiah, Jared~D Kaplan, Prafulla Dhariwal, Arvind Neelakantan, Pranav Shyam, Girish Sastry, Amanda Askell, et~al.
\newblock Language models are few-shot learners.
\newblock {\em Advances in neural information processing systems}, 33:1877--1901, 2020.

\bibitem{cao2022augmented}
Jie Cao and Jing Xiao.
\newblock An augmented benchmark dataset for geometric question answering through dual parallel text encoding.
\newblock In {\em Proceedings of the 29th International Conference on Computational Linguistics}, pages 1511--1520, 2022.

\bibitem{chen2025sharegpt4v}
Lin Chen, Jinsong Li, Xiaoyi Dong, Pan Zhang, Conghui He, Jiaqi Wang, Feng Zhao, and Dahua Lin.
\newblock Sharegpt4v: Improving large multi-modal models with better captions.
\newblock In {\em European Conference on Computer Vision}, pages 370--387. Springer, 2025.

\bibitem{mmstar}
Lin Chen, Jinsong Li, Xiaoyi Dong, Pan Zhang, Yuhang Zang, Zehui Chen, Haodong Duan, Jiaqi Wang, Yu~Qiao, Dahua Lin, et~al.
\newblock Are we on the right way for evaluating large vision-language models?
\newblock {\em arXiv preprint arXiv:2403.20330}, 2024.

\bibitem{chen2024internvl}
Zhe Chen, Jiannan Wu, Wenhai Wang, Weijie Su, Guo Chen, Sen Xing, Muyan Zhong, Qinglong Zhang, Xizhou Zhu, Lewei Lu, et~al.
\newblock Internvl: Scaling up vision foundation models and aligning for generic visual-linguistic tasks.
\newblock In {\em Proceedings of the IEEE/CVF Conference on Computer Vision and Pattern Recognition}, pages 24185--24198, 2024.

\bibitem{chiang-chen-2019-semantically}
Ting-Rui Chiang and Yun-Nung Chen.
\newblock Semantically-aligned equation generation for solving and reasoning math word problems.
\newblock In Jill Burstein, Christy Doran, and Thamar Solorio, editors, {\em Proceedings of the 2019 Conference of the North {A}merican Chapter of the Association for Computational Linguistics: Human Language Technologies, Volume 1 (Long and Short Papers)}, pages 2656--2668, Minneapolis, Minnesota, June 2019. Association for Computational Linguistics.

\bibitem{chowdhery2023palm}
Aakanksha Chowdhery, Sharan Narang, Jacob Devlin, Maarten Bosma, Gaurav Mishra, Adam Roberts, Paul Barham, Hyung~Won Chung, Charles Sutton, Sebastian Gehrmann, et~al.
\newblock Palm: Scaling language modeling with pathways.
\newblock {\em Journal of Machine Learning Research}, 24(240):1--113, 2023.

\bibitem{deitke2024molmo}
Matt Deitke, Christopher Clark, Sangho Lee, Rohun Tripathi, Yue Yang, Jae~Sung Park, Mohammadreza Salehi, Niklas Muennighoff, Kyle Lo, Luca Soldaini, et~al.
\newblock Molmo and pixmo: Open weights and open data for state-of-the-art multimodal models.
\newblock {\em arXiv preprint arXiv:2409.17146}, 2024.

\bibitem{vlmevalkit}
Haodong Duan, Junming Yang, Yuxuan Qiao, Xinyu Fang, Lin Chen, Yuan Liu, Xiaoyi Dong, Yuhang Zang, Pan Zhang, Jiaqi Wang, Dahua Lin, and Kai Chen.
\newblock Vlmevalkit: An open-source toolkit for evaluating large multi-modality models, 2024.

\bibitem{fu2024blink}
Xingyu Fu, Yushi Hu, Bangzheng Li, Yu~Feng, Haoyu Wang, Xudong Lin, Dan Roth, Noah~A Smith, Wei-Chiu Ma, and Ranjay Krishna.
\newblock Blink: Multimodal large language models can see but not perceive.
\newblock {\em arXiv preprint arXiv:2404.12390}, 2024.

\bibitem{gao2023g}
Jiahui Gao, Renjie Pi, Jipeng Zhang, Jiacheng Ye, Wanjun Zhong, Yufei Wang, Lanqing Hong, Jianhua Han, Hang Xu, Zhenguo Li, et~al.
\newblock G-llava: Solving geometric problem with multi-modal large language model.
\newblock {\em arXiv preprint arXiv:2312.11370}, 2023.

\bibitem{garcez2019neural}
Artur~d'Avila Garcez, Marco Gori, Luis~C Lamb, Luciano Serafini, Michael Spranger, and Son~N Tran.
\newblock Neural-symbolic computing: An effective methodology for principled integration of machine learning and reasoning.
\newblock {\em arXiv preprint arXiv:1905.06088}, 2019.

\bibitem{geva2021did}
Mor Geva, Daniel Khashabi, Elad Segal, Tushar Khot, Dan Roth, and Jonathan Berant.
\newblock Did aristotle use a laptop? a question answering benchmark with implicit reasoning strategies.
\newblock {\em Transactions of the Association for Computational Linguistics}, 9:346--361, 2021.

\bibitem{golovnevaroscoe}
Olga Golovneva, Moya~Peng Chen, Spencer Poff, Martin Corredor, Luke Zettlemoyer, Maryam Fazel-Zarandi, and Asli Celikyilmaz.
\newblock Roscoe: A suite of metrics for scoring step-by-step reasoning.
\newblock In {\em The Eleventh International Conference on Learning Representations}.

\bibitem{goyal2022news}
Tanya Goyal, Junyi~Jessy Li, and Greg Durrett.
\newblock News summarization and evaluation in the era of gpt-3.
\newblock {\em arXiv preprint arXiv:2209.12356}, 2022.

\bibitem{hallusionbench}
Tianrui Guan, Fuxiao Liu, Xiyang Wu, Ruiqi Xian, Zongxia Li, Xiaoyu Liu, Xijun Wang, Lichang Chen, Furong Huang, Yaser Yacoob, Dinesh Manocha, and Tianyi Zhou.
\newblock Hallusionbench: An advanced diagnostic suite for entangled language hallucination and visual illusion in large vision-language models.
\newblock In {\em Proceedings of the IEEE/CVF Conference on Computer Vision and Pattern Recognition (CVPR)}, pages 14375--14385, June 2024.

\bibitem{gupta2023visual}
Tanmay Gupta and Aniruddha Kembhavi.
\newblock Visual programming: Compositional visual reasoning without training.
\newblock In {\em Proceedings of the IEEE/CVF Conference on Computer Vision and Pattern Recognition}, pages 14953--14962, 2023.

\bibitem{hu2023look}
Yingdong Hu, Fanqi Lin, Tong Zhang, Li~Yi, and Yang Gao.
\newblock Look before you leap: Unveiling the power of gpt-4v in robotic vision-language planning.
\newblock {\em arXiv preprint arXiv:2311.17842}, 2023.

\bibitem{huang2022large}
Jiaxin Huang, Shixiang~Shane Gu, Le~Hou, Yuexin Wu, Xuezhi Wang, Hongkun Yu, and Jiawei Han.
\newblock Large language models can self-improve.
\newblock {\em arXiv preprint arXiv:2210.11610}, 2022.

\bibitem{johnson2017clevr}
Justin Johnson, Bharath Hariharan, Laurens Van Der~Maaten, Li~Fei-Fei, C~Lawrence~Zitnick, and Ross Girshick.
\newblock Clevr: A diagnostic dataset for compositional language and elementary visual reasoning.
\newblock In {\em Proceedings of the IEEE conference on computer vision and pattern recognition}, pages 2901--2910, 2017.

\bibitem{johnson2016densecap}
Justin Johnson, Andrej Karpathy, and Li~Fei-Fei.
\newblock Densecap: Fully convolutional localization networks for dense captioning.
\newblock In {\em Proceedings of the IEEE conference on computer vision and pattern recognition}, pages 4565--4574, 2016.

\bibitem{junnan2017people}
Li~Junnan, Wong~Yong Kang, Zhao Qi, and Mohan~S Kankanhalli.
\newblock People in social context (pisc) dataset.
\newblock 2017.

\bibitem{kembhavi2016diagram}
Aniruddha Kembhavi, Mike Salvato, Eric Kolve, Minjoon Seo, Hannaneh Hajishirzi, and Ali Farhadi.
\newblock A diagram is worth a dozen images.
\newblock In {\em Computer Vision--ECCV 2016: 14th European Conference, Amsterdam, The Netherlands, October 11--14, 2016, Proceedings, Part IV 14}, pages 235--251. Springer, 2016.

\bibitem{ai2d}
Aniruddha Kembhavi, Mike Salvato, Eric Kolve, Minjoon Seo, Hannaneh Hajishirzi, and Ali Farhadi.
\newblock A diagram is worth a dozen images.
\newblock In {\em Computer Vision--ECCV 2016: 14th European Conference, Amsterdam, The Netherlands, October 11--14, 2016, Proceedings, Part IV 14}, pages 235--251. Springer, 2016.

\bibitem{kembhavi2017you}
Aniruddha Kembhavi, Minjoon Seo, Dustin Schwenk, Jonghyun Choi, Ali Farhadi, and Hannaneh Hajishirzi.
\newblock Are you smarter than a sixth grader? textbook question answering for multimodal machine comprehension.
\newblock In {\em Proceedings of the IEEE Conference on Computer Vision and Pattern recognition}, pages 4999--5007, 2017.

\bibitem{khan2014geometry}
Salman~Hameed Khan, Mohammed Bennamoun, Ferdous Sohel, and Roberto Togneri.
\newblock Geometry driven semantic labeling of indoor scenes.
\newblock In {\em Computer Vision--ECCV 2014: 13th European Conference, Zurich, Switzerland, September 6-12, 2014, Proceedings, Part I 13}, pages 679--694. Springer, 2014.

\bibitem{lampinen2022tell}
Andrew~K Lampinen, Nicholas Roy, Ishita Dasgupta, Stephanie~CY Chan, Allison Tam, James Mcclelland, Chen Yan, Adam Santoro, Neil~C Rabinowitz, Jane Wang, et~al.
\newblock Tell me why! explanations support learning relational and causal structure.
\newblock In {\em International Conference on Machine Learning}, pages 11868--11890. PMLR, 2022.

\bibitem{le2023bloom}
Teven Le~Scao, Angela Fan, Christopher Akiki, Ellie Pavlick, Suzana Ili{\'c}, Daniel Hesslow, Roman Castagn{\'e}, Alexandra~Sasha Luccioni, Fran{\c{c}}ois Yvon, Matthias Gall{\'e}, et~al.
\newblock Bloom: A 176b-parameter open-access multilingual language model.
\newblock 2023.

\bibitem{lindstrom2022clevr}
Adam~Dahlgren Lindstr{\"o}m and Savitha~Sam Abraham.
\newblock Clevr-math: A dataset for compositional language, visual and mathematical reasoning.
\newblock {\em arXiv preprint arXiv:2208.05358}, 2022.

\bibitem{mmbench}
Yuan Liu, Haodong Duan, Yuanhan Zhang, Bo~Li, Songyang Zhang, Wangbo Zhao, Yike Yuan, Jiaqi Wang, Conghui He, Ziwei Liu, et~al.
\newblock Mmbench: Is your multi-modal model an all-around player?
\newblock In {\em European Conference on Computer Vision}, pages 216--233. Springer, 2025.

\bibitem{liu2024nvila}
Zhijian Liu, Ligeng Zhu, Baifeng Shi, Zhuoyang Zhang, Yuming Lou, Shang Yang, Haocheng Xi, Shiyi Cao, Yuxian Gu, Dacheng Li, et~al.
\newblock Nvila: Efficient frontier visual language models.
\newblock {\em arXiv preprint arXiv:2412.04468}, 2024.

\bibitem{lu2023mathvista}
Pan Lu, Hritik Bansal, Tony Xia, Jiacheng Liu, Chunyuan Li, Hannaneh Hajishirzi, Hao Cheng, Kai-Wei Chang, Michel Galley, and Jianfeng Gao.
\newblock Mathvista: Evaluating mathematical reasoning of foundation models in visual contexts.
\newblock {\em arXiv preprint arXiv:2310.02255}, 2023.

\bibitem{lu2024mathvista}
Pan Lu, Hritik Bansal, Tony Xia, Jiacheng Liu, Chunyuan Li, Hannaneh Hajishirzi, Hao Cheng, Kai-Wei Chang, Michel Galley, and Jianfeng Gao.
\newblock Mathvista: Evaluating mathematical reasoning of foundation models in visual contexts.
\newblock In {\em International Conference on Learning Representations (ICLR)}, 2024.

\bibitem{mathvista}
Pan Lu, Hritik Bansal, Tony Xia, Jiacheng Liu, Chunyuan Li, Hannaneh Hajishirzi, Hao Cheng, Kai-Wei Chang, Michel Galley, and Jianfeng Gao.
\newblock Mathvista: Evaluating mathematical reasoning of foundation models in visual contexts.
\newblock In {\em International Conference on Learning Representations (ICLR)}, 2024.

\bibitem{lu2022learn}
Pan Lu, Swaroop Mishra, Tanglin Xia, Liang Qiu, Kai-Wei Chang, Song-Chun Zhu, Oyvind Tafjord, Peter Clark, and Ashwin Kalyan.
\newblock Learn to explain: Multimodal reasoning via thought chains for science question answering.
\newblock {\em Advances in Neural Information Processing Systems}, 35:2507--2521, 2022.

\bibitem{lu2024ovis}
Shiyin Lu, Yang Li, Qing-Guo Chen, Zhao Xu, Weihua Luo, Kaifu Zhang, and Han-Jia Ye.
\newblock Ovis: Structural embedding alignment for multimodal large language model.
\newblock {\em arXiv:2405.20797}, 2024.

\bibitem{lu2024llamax}
Yinquan Lu, Wenhao Zhu, Lei Li, Yu~Qiao, and Fei Yuan.
\newblock Llamax: Scaling linguistic horizons of llm by enhancing translation capabilities beyond 100 languages.
\newblock {\em arXiv preprint arXiv:2407.05975}, 2024.

\bibitem{madaan2024self}
Aman Madaan, Niket Tandon, Prakhar Gupta, Skyler Hallinan, Luyu Gao, Sarah Wiegreffe, Uri Alon, Nouha Dziri, Shrimai Prabhumoye, Yiming Yang, et~al.
\newblock Self-refine: Iterative refinement with self-feedback.
\newblock {\em Advances in Neural Information Processing Systems}, 36, 2024.

\bibitem{masry2022chartqa}
Ahmed Masry, Do~Xuan Long, Jia~Qing Tan, Shafiq Joty, and Enamul Hoque.
\newblock Chartqa: A benchmark for question answering about charts with visual and logical reasoning.
\newblock {\em arXiv preprint arXiv:2203.10244}, 2022.

\bibitem{mathew2021docvqa}
Minesh Mathew, Dimosthenis Karatzas, and CV~Jawahar.
\newblock Docvqa: A dataset for vqa on document images.
\newblock In {\em Proceedings of the IEEE/CVF winter conference on applications of computer vision}, pages 2200--2209, 2021.

\bibitem{mathew2020docvqa}
Minesh Mathew, Dimosthenis Karatzas, R~Manmatha, and CV~Jawahar.
\newblock Docvqa: A dataset for vqa on document images. corr abs/2007.00398 (2020).
\newblock {\em arXiv preprint arXiv:2007.00398}, 2020.

\bibitem{llama3.2}
{Meta AI}.
\newblock Llama 3.2: Revolutionizing edge ai and vision with open, customizable models.
\newblock \url{https://ai.meta.com/blog/llama-3-2-connect-2024-vision-edge-mobile-devices/}, 2024.

\bibitem{gpt4omini}
{OpenAI}.
\newblock Gpt-4o mini: advancing cost-efficient intelligence.
\newblock \url{https://openai.com/index/gpt-4o-mini-advancing-cost-efficient-intelligence/}, 2024.

\bibitem{openai2024o1}
OpenAI.
\newblock Introducing openai o1-preview, 2024.
\newblock Accessed: 2024-12-16.

\bibitem{prasad2023receval}
Archiki Prasad, Swarnadeep Saha, Xiang Zhou, and Mohit Bansal.
\newblock Receval: Evaluating reasoning chains via correctness and informativeness.
\newblock In {\em The 2023 Conference on Empirical Methods in Natural Language Processing}.

\bibitem{rafailov2024direct}
Rafael Rafailov, Archit Sharma, Eric Mitchell, Christopher~D Manning, Stefano Ermon, and Chelsea Finn.
\newblock Direct preference optimization: Your language model is secretly a reward model.
\newblock {\em Advances in Neural Information Processing Systems}, 36, 2024.

\bibitem{gemini1.5}
Machel Reid, Nikolay Savinov, Denis Teplyashin, Dmitry Lepikhin, Timothy Lillicrap, Jean baptiste Alayrac, Radu Soricut, Angeliki Lazaridou, Orhan Firat, Julian Schrittwieser, and et~al.
\newblock Gemini 1.5: Unlocking multimodal understanding across millions of tokens of context, 2024.

\bibitem{roy-roth-2015-solving}
Subhro Roy and Dan Roth.
\newblock Solving general arithmetic word problems.
\newblock In Llu{\'\i}s M{\`a}rquez, Chris Callison-Burch, and Jian Su, editors, {\em Proceedings of the 2015 Conference on Empirical Methods in Natural Language Processing}, pages 1743--1752, Lisbon, Portugal, September 2015. Association for Computational Linguistics.

\bibitem{schwenk2022okvqa}
Dustin Schwenk, Apoorv Khandelwal, Christopher Clark, Kenneth Marino, and Roozbeh Mottaghi.
\newblock A-okvqa: A benchmark for visual question answering using world knowledge.
\newblock In {\em European conference on computer vision}, pages 146--162. Springer, 2022.

\bibitem{touvron2023llama}
Hugo Touvron, Thibaut Lavril, Gautier Izacard, Xavier Martinet, Marie-Anne Lachaux, Timoth{\'e}e Lacroix, Baptiste Rozi{\`e}re, Naman Goyal, Eric Hambro, Faisal Azhar, et~al.
\newblock Llama: Open and efficient foundation language models.
\newblock {\em arXiv preprint arXiv:2302.13971}, 2023.

\bibitem{touvron2023llama2}
Hugo Touvron, Louis Martin, Kevin Stone, Peter Albert, Amjad Almahairi, Yasmine Babaei, Nikolay Bashlykov, Soumya Batra, Prajjwal Bhargava, Shruti Bhosale, et~al.
\newblock Llama 2: Open foundation and fine-tuned chat models.
\newblock {\em arXiv preprint arXiv:2307.09288}, 2023.

\bibitem{vayani2024alm}
Ashmal Vayani, Dinura Dissanayake, Hasindri Watawana, Noor Ahsan, Nevasini Sasikumar, Omkar Thawakar, Henok~Biadglign Ademtew, Yahya Hmaiti, Amandeep Kumar, Kartik Kuckreja, Mykola Maslych, Wafa~Al Ghallabi, Mihail Mihaylov, Chao Qin, Abdelrahman~M Shaker, Mike Zhang, Mahardika~Krisna Ihsani, Amiel Esplana, Monil Gokani, Shachar Mirkin, Harsh Singh, Ashay Srivastava, Endre Hamerlik, Fathinah~Asma Izzati, Fadillah~Adamsyah Maani, Sebastian Cavada, Jenny Chim, Rohit Gupta, Sanjay Manjunath, Kamila Zhumakhanova, Feno~Heriniaina Rabevohitra, Azril Amirudin, Muhammad Ridzuan, Daniya Kareem, Ketan More, Kunyang Li, Pramesh Shakya, Muhammad Saad, Amirpouya Ghasemaghaei, Amirbek Djanibekov, Dilshod Azizov, Branislava Jankovic, Naman Bhatia, Alvaro Cabrera, Johan Obando-Ceron, Olympiah Otieno, Fabian Farestam, Muztoba Rabbani, Sanoojan Baliah, Santosh Sanjeev, Abduragim Shtanchaev, Maheen Fatima, Thao Nguyen, Amrin Kareem, Toluwani Aremu, Nathan Xavier, Amit Bhatkal, Hawau Toyin, Aman Chadha, Hisham Cholakkal,
  Rao~Muhammad Anwer, Michael Felsberg, Jorma Laaksonen, Thamar Solorio, Monojit Choudhury, Ivan Laptev, Mubarak Shah, Salman Khan, and Fahad Khan.
\newblock All languages matter: Evaluating lmms on culturally diverse 100 languages, 2024.

\bibitem{vedantam2019probabilistic}
Ramakrishna Vedantam, Karan Desai, Stefan Lee, Marcus Rohrbach, Dhruv Batra, and Devi Parikh.
\newblock Probabilistic neural symbolic models for interpretable visual question answering.
\newblock In {\em International Conference on Machine Learning}, pages 6428--6437. PMLR, 2019.

\bibitem{wang2023selfconsistency}
Xuezhi Wang, Jason Wei, Dale Schuurmans, Quoc~V Le, Ed~H. Chi, Sharan Narang, Aakanksha Chowdhery, and Denny Zhou.
\newblock Self-consistency improves chain of thought reasoning in language models.
\newblock In {\em The Eleventh International Conference on Learning Representations}, 2023.

\bibitem{wang2023element}
Yiming Wang, Zhuosheng Zhang, and Rui Wang.
\newblock Element-aware summarization with large language models: Expert-aligned evaluation and chain-of-thought method.
\newblock {\em arXiv preprint arXiv:2305.13412}, 2023.

\bibitem{wei2022chain}
Jason Wei, Xuezhi Wang, Dale Schuurmans, Maarten Bosma, Fei Xia, Ed~Chi, Quoc~V Le, Denny Zhou, et~al.
\newblock Chain-of-thought prompting elicits reasoning in large language models.
\newblock {\em Advances in neural information processing systems}, 35:24824--24837, 2022.

\bibitem{weng2022large}
Yixuan Weng, Minjun Zhu, Fei Xia, Bin Li, Shizhu He, Shengping Liu, Bin Sun, Kang Liu, and Jun Zhao.
\newblock Large language models are better reasoners with self-verification.
\newblock {\em arXiv preprint arXiv:2212.09561}, 2022.

\bibitem{wu2024comparative}
Siwei Wu, Zhongyuan Peng, Xinrun Du, Tuney Zheng, Minghao Liu, Jialong Wu, Jiachen Ma, Yizhi Li, Jian Yang, Wangchunshu Zhou, et~al.
\newblock A comparative study on reasoning patterns of openai's o1 model.
\newblock {\em arXiv preprint arXiv:2410.13639}, 2024.

\bibitem{xiao2024logicvistamultimodalllmlogical}
Yijia Xiao, Edward Sun, Tianyu Liu, and Wei Wang.
\newblock Logicvista: Multimodal llm logical reasoning benchmark in visual contexts, 2024.

\bibitem{xiong2024llava}
Tianyi Xiong, Xiyao Wang, Dong Guo, Qinghao Ye, Haoqi Fan, Quanquan Gu, Heng Huang, and Chunyuan Li.
\newblock Llava-critic: Learning to evaluate multimodal models.
\newblock {\em arXiv preprint arXiv:2410.02712}, 2024.

\bibitem{llava_cot}
Guowei Xu, Peng Jin, Li~Hao, Yibing Song, Lichao Sun, and Li~Yuan.
\newblock Llava-o1: Let vision language models reason step-by-step.
\newblock {\em arXiv preprint arXiv:2411.10440}, 2024.

\bibitem{xu2024llava}
Guowei Xu, Peng Jin, Li~Hao, Yibing Song, Lichao Sun, and Li~Yuan.
\newblock Llava-o1: Let vision language models reason step-by-step.
\newblock {\em arXiv preprint arXiv:2411.10440}, 2024.

\bibitem{yao2024mulberry}
Huanjin Yao, Jiaxing Huang, Wenhao Wu, Jingyi Zhang, Yibo Wang, Shunyu Liu, Yingjie Wang, Yuxin Song, Haocheng Feng, Li~Shen, et~al.
\newblock Mulberry: Empowering mllm with o1-like reasoning and reflection via collective monte carlo tree search.
\newblock {\em arXiv preprint arXiv:2412.18319}, 2024.

\bibitem{yao2024tree}
Shunyu Yao, Dian Yu, Jeffrey Zhao, Izhak Shafran, Tom Griffiths, Yuan Cao, and Karthik Narasimhan.
\newblock Tree of thoughts: Deliberate problem solving with large language models.
\newblock {\em Advances in Neural Information Processing Systems}, 36, 2024.

\bibitem{yao2024minicpm}
Yuan Yao, Tianyu Yu, Ao~Zhang, Chongyi Wang, Junbo Cui, Hongji Zhu, Tianchi Cai, Haoyu Li, Weilin Zhao, Zhihui He, et~al.
\newblock Minicpm-v: A gpt-4v level mllm on your phone.
\newblock {\em arXiv preprint arXiv:2408.01800}, 2024.

\bibitem{mmvet}
Weihao Yu, Zhengyuan Yang, Linjie Li, Jianfeng Wang, Kevin Lin, Zicheng Liu, Xinchao Wang, and Lijuan Wang.
\newblock Mm-vet: Evaluating large multimodal models for integrated capabilities.
\newblock {\em arXiv preprint arXiv:2308.02490}, 2023.

\bibitem{yue2023mmmu}
Xiang Yue, Yuansheng Ni, Kai Zhang, Tianyu Zheng, Ruoqi Liu, Ge~Zhang, Samuel Stevens, Dongfu Jiang, Weiming Ren, Yuxuan Sun, Cong Wei, Botao Yu, Ruibin Yuan, Renliang Sun, Ming Yin, Boyuan Zheng, Zhenzhu Yang, Yibo Liu, Wenhao Huang, Huan Sun, Yu~Su, and Wenhu Chen.
\newblock Mmmu: A massive multi-discipline multimodal understanding and reasoning benchmark for expert agi.
\newblock In {\em Proceedings of CVPR}, 2024.

\bibitem{zhang2024improve}
Ruohong Zhang, Bowen Zhang, Yanghao Li, Haotian Zhang, Zhiqing Sun, Zhe Gan, Yinfei Yang, Ruoming Pang, and Yiming Yang.
\newblock Improve vision language model chain-of-thought reasoning.
\newblock {\em arXiv preprint arXiv:2410.16198}, 2024.

\bibitem{zhang2022opt}
Susan Zhang, Stephen Roller, Naman Goyal, Mikel Artetxe, Moya Chen, Shuohui Chen, Christopher Dewan, Mona Diab, Xian Li, Xi~Victoria Lin, et~al.
\newblock Opt: Open pre-trained transformer language models.
\newblock {\em arXiv preprint arXiv:2205.01068}, 2022.

\bibitem{zhang2024benchmarking}
Tianyi Zhang, Faisal Ladhak, Esin Durmus, Percy Liang, Kathleen McKeown, and Tatsunori~B Hashimoto.
\newblock Benchmarking large language models for news summarization.
\newblock {\em Transactions of the Association for Computational Linguistics}, 12:39--57, 2024.

\bibitem{zou2024dynamic}
Chengke Zou, Xingang Guo, Rui Yang, Junyu Zhang, Bin Hu, and Huan Zhang.
\newblock Dynamath: A dynamic visual benchmark for evaluating mathematical reasoning robustness of vision language models, 2024.

\end{thebibliography}


\newpage
\appendix

\section{Appendix}

\subsection{Generating reasoning Steps from Closed Sourced Models} 
\label{appendix:generating_reasoning_steps_from_target_models}

We designed a structured system prompt to guide closed-source models like GPT-4o~\cite{openai2024gpt4ocard}, Claude~\cite{claude3.5sonnet_blog}, and Gemini~\cite{gemini1.5} in generating detailed, step-by-step reasoning for complex tasks. The prompt requires the model to describe each action to be taken and explain how it is executed, ensuring a clear and logical progression throughout the reasoning process. To account for varying levels of complexity, the prompt allows the model to take as many steps as necessary, ensuring that the solution is systematically derived. Additionally, the prompt emphasizes the use of visual elements, guiding the model to reference provided images or diagrams explicitly in its reasoning steps.

The prompt is further designed to handle ambiguity effectively by instructing the model to respond with "None of the choices provided" when no valid options are available. This ensures robustness and prevents the generation of forced or inaccurate conclusions. By enforcing a logical flow, grounding the reasoning in visual inputs, and providing explicit instructions for ambiguous scenarios, this prompt enables consistent, interpretable, and reliable reasoning outputs across various multimodal tasks.

\begin{tcolorbox}[colback=gray!5, colframe=black!45,
    fonttitle=\bfseries, title= System Prompt used for the generation of reasoning steps
    ]
\small

\lstset{breaklines=true, basicstyle=\small\ttfamily} 

\begin{lstlisting}
When answering the question based on the provided image(s), 
follow a structured reasoning process and provide the final answer 
after solving it step by step. Use the following format for 
your response:

Step-by-Step Process:
    Step 1: Describe the action to be taken.
    Action 1: Explain the execution of the first action.
    
    Step 2: Describe the next action to be taken.
    Action 2: Explain the execution of the second action.
    
    Step 3: Describe the next action to be taken.
    Action 3: Explain the execution of the second action.
    
    ...continue as needed...take as many steps you want.
    
    Step n: Describe the final action to be taken.
    Action n: Execute the final action leading to the conclusion.
    
    Final Answer: Provide the final solution or conclusion derived from the reasoning process.

Ensure each step logically follows the previous one, and explicitly 
detail how the image(s) guide the solution at every stage. Also if 
options are present and none of options are correct.
Please response None of the choices provided. 

\end{lstlisting}

\end{tcolorbox}

\newpage

\subsection{System Prompt used to Evaluate Reasoning Steps } 
\label{appendix:system_prompt_for_reasoning_eval}
The following system prompt was used to evaluate the reasoning steps of the target model. It defines a structured framework to assess the alignment, coherence, and quality of reasoning through multiple metrics, including faithfulness, informativeness, repetition, hallucination, redundancy, semantic coverage, reasoning alignment, commonsense, and completeness of steps. Each metric is scored on a scale of 1-10, with detailed guidelines ensuring consistent and objective evaluations.

\begin{tcolorbox}[colback=gray!5, colframe=black!45,
    fonttitle=\bfseries, title=System Prompt used to evaluate the reasoning steps
    ]
\small

\lstset{breaklines=true, basicstyle=\small\ttfamily} 

\begin{lstlisting}
You are a reasoning evaluator designed to assess the alignment, coherence, and quality of reasoning steps in text responses. Your task is to evaluate reasoning steps between the *ground truth* and the *LLM response* using the following metrics:

1. **Faithfulness-Step (1-10):**  
   - Definition: Measures how well the reasoning steps in the LLM response align with the source reasoning steps.
   - Scoring Guidelines:  
     - 9-10: All or almost all steps match or closely reflect the ground truth reasoning.  
     - 7-8: Most steps are aligned, with minor deviations.  
     - 5-6: Some steps align, but several are missing or significantly altered.  
     - 3-4: Few steps align correctly; most are off or missing.  
     - 1-2: The majority of steps are not aligned with the source.

2. **Faithfulness-Token (1-10):**  
   - Definition: Extends Faithfulness-Step to a token-level granularity, checking if the content within each reasoning step is true to the source.
   - Scoring Guidelines:  
     - 9-10: Token-level details mirror the ground truth closely.  
     - 7-8: Minor token-level deviations but largely faithful.  
     - 5-6: Noticeable inaccuracies in token-level details.  
     - 3-4: Several token-level discrepancies.  
     - 1-2: Most token-level details are incorrect or fabricated.

3. **Informativeness-Step (Info-Step) (1-10):**  
   - Definition: Measures how well the reasoning steps extract all relevant information from the source.
   - Scoring Guidelines:  
     - 9-10: Almost all critical information steps are present and accurate.  
     - 7-8: Most important points are included, with minor omissions.  
     - 5-6: Some key information is missing or underdeveloped.  
     - 3-4: Limited inclusion of critical content.  
     - 1-2: Very poor extraction of relevant information.

4. **Repetition-Token (1-10):**  
   - Definition: Identifies repeated or unnecessarily paraphrased reasoning steps within the hypothesis.
   - Scoring Guidelines:  
     - 9-10: No or minimal unnecessary repetition.  
     - 7-8: Minor repetition that doesn't impede clarity.  
     - 5-6: Noticeable repetition that doesn't add value.  
     - 3-4: Frequent repetition that disrupts coherence.  
     - 1-2: Excessive repetition reducing the quality of reasoning.
     
\end{lstlisting}

\end{tcolorbox}

\begin{tcolorbox}[colback=gray!5, colframe=black!45,
    fonttitle=\bfseries, title=System Prompt used to evaluate the reasoning steps continued...
    ]
\small

\lstset{breaklines=true, basicstyle=\small\ttfamily} 

\begin{lstlisting}
5. **Hallucination (1-10):**  
   - Definition: Detect irrelevant or invented reasoning steps not aligned with the source.
   - Scoring Guidelines:  
     - 9-10: No hallucinations; all reasoning is grounded in the source.  
     - 7-8: One or two minor hallucinations.  
     - 5-6: Several steps contain invented or irrelevant details.  
     - 3-4: Many hallucinations, but some grounding remains.  
     - 1-2: Mostly hallucinated reasoning.

6. **Redundancy (1-10):**  
   - Definition: Identify redundant reasoning steps that do not add value.
   - Scoring Guidelines:  
     - 9-10: No unnecessary steps; very concise.  
     - 7-8: Minor redundancy.  
     - 5-6: Some steps clearly unnecessary.  
     - 3-4: Many redundant steps.  
     - 1-2: Excessive redundancy that hampers clarity.

7. **Semantic Coverage-Step (1-10):**  
   - Definition: How well the hypothesis covers the essential semantic elements from the source reasoning steps.
   - Scoring Guidelines:  
     - 9-10: Almost complete semantic coverage of all important elements.  
     - 7-8: Good coverage but some minor elements are missing.  
     - 5-6: Partial coverage with noticeable gaps.  
     - 3-4: Significant semantic gaps.  
     - 1-2: Very poor coverage of essential meaning.

8. **Reasoning Alignment (1-10):**  
   - Definition: Overall alignment between the hypothesis and the reference reasoning chain.
   - Scoring Guidelines:  
     - 9-10: Very closely aligned, minimal divergence.  
     - 7-8: Mostly aligned, with some minor issues.  
     - 5-6: Some alignment, but also several misalignments.  
     - 3-4: Poor alignment, though occasional matches.  
     - 1-2: Fundamentally misaligned reasoning.

9. **Commonsense (1-10):**  
   - Definition: Check for missing commonsense reasoning required to solve the problem.
   - Scoring Guidelines:  
     - 9-10: Adequate commonsense reasoning present.  
     - 7-8: Minor commonsense gaps but mostly adequate.  
     - 5-6: Noticeable commonsense gaps.  
     - 3-4: Many commonsense steps missing.  
     - 1-2: Almost entirely lacking necessary commonsense.

\end{lstlisting}

\end{tcolorbox}

\begin{tcolorbox}[colback=gray!5, colframe=black!45,
    fonttitle=\bfseries, title=System Prompt used to evaluate the reasoning steps continued...
    ]
\small

\lstset{breaklines=true, basicstyle=\small\ttfamily} 

\begin{lstlisting}
10. **Missing Step (1-10):**  
    - Definition: Identify if any necessary reasoning steps are missing.
    - Scoring Guidelines:  
      - 9-10: No critical steps missing.  
      - 7-8: Minor missing steps that don't significantly affect the conclusion.  
      - 5-6: Some important steps absent, affecting the outcome.  
      - 3-4: Several crucial missing steps.  
      - 1-2: Major gaps; the reasoning chain is incomplete.

**Additional Instructions for Consistency:**

- Always follow the above scoring guidelines strictly.  
- Before scoring, re-read both the ground truth and the LLM response carefully.  
- Compare the reasoning steps directly to determine where they align or diverge.
- Use the provided scoring benchmarks (anchor examples, if any) as a reference to maintain consistency across evaluations.
- Avoid subjective interpretation and adhere to the given thresholds.
- Once scores for all metrics are determined, compute the Overall Score as the average of all metric scores.
- Provide the final output as a Python dictionary with the structure only don't add a anything extra , beacuase your out will be used in code pipeline. So single change in you output will crash whole system. :

# Example output : {'Faithfulness-Step': 8.0, 'Faithfulness-Token': 7.5, 'Informativeness-Step': 8.5, 'Repetition-Token': 9.0, 'Hallucination': 9.5, 'Redundancy': 8.0, 'Semantic Coverage-Step': 8.5, 'Reasoning Alignment': 8.0, 'Commonsense': 9.0, 'Missing Step': 8.5 , 'Overall Score': 8.65}

# Do not give output in following format :

```python
{
  'Faithfulness-Step': 1.0,
  'Faithfulness-Token': 1.0,
  'Informativeness-Step': 1.0,
  'Repetition-Token': 9.0,
  'Hallucination': 1.0,
  'Redundancy': 9.0,
  'Semantic Coverage-Step': 1.0,
  'Reasoning Alignment': 1.0,
  'Commonsense': 1.0,
  'Missing Step': 1.0,
  'Overall Score': 2.6
}
```
\end{lstlisting}

\end{tcolorbox}

\newpage

\subsection{Response format used to generate structured evaluation scores } 
\label{appendix:response_format_to_generate_structured_evaluation_scores}

To further ensure the evaluation framework generates consistent and interpretable outputs, we designed the response format using a well-defined JSON schema. This schema serves as a blueprint, enforcing strict adherence to a structured format while capturing detailed scores for each metric in a systematic and transparent manner. By standardizing the output structure, the schema facilitates easier comparison between models, reduces ambiguity, and enhances the reproducibility of results.

The JSON schema is carefully tailored to accommodate the unique aspects of our evaluation process, such as step-by-step reasoning, metric-specific scores, and logical flow validation. Each response is divided into key components, including reasoning steps, metric scores, and final answers, ensuring that all critical aspects of the model's performance are systematically captured. This level of detail not only improves interpretability but also enables fine-grained analysis of strengths and weaknesses across models.

Additionally, the schema supports modularity, allowing seamless integration of new metrics or evaluation criteria as the benchmark evolves. By adopting this structured approach, we ensure that the evaluation framework remains robust, scalable, and adaptable to future advancements in multimodal reasoning research.

\begin{tcolorbox}[colback=gray!5, colframe=black!45,
    fonttitle=\bfseries, title=Response format provided to LLM's which supports structured-output
    ]
\small
\lstset{breaklines=true, basicstyle=\small\ttfamily} 
\begin{lstlisting}
response_format = {
    "type": "json_schema",
    "json_schema": {
        "name": "EvaluationScores",
        "strict": True,
        "schema": {
            "type": "object",
            "properties": {
                "Faithfulness-Step": {"type": "number"},
                "Faithfulness-Token": {"type": "number"},
                "Informativeness-Step": {"type": "number"},
                "Repetition-Token": {"type": "number"},
                "Hallucination": {"type": "number"},
                "Redundancy": {"type": "number"},
                "Semantic Coverage-Step": {"type": "number"},
                "Reasoning Alignment": {"type": "number"},
                "Commonsense": {"type": "number"},
                "Missing Step": {"type": "number"},
                "Overall Score": {"type": "number"}
            },
            "required": [
                "Faithfulness-Step",
                "Faithfulness-Token",
                "Informativeness-Step",
                "Repetition-Token",
                "Hallucination",
                "Redundancy",
                "Semantic Coverage-Step",
                "Reasoning Alignment",
                "Commonsense",
                "Missing Step",
                "Overall Score"
            ],
            "additionalProperties": False
        }
    }
}

\end{lstlisting}

\end{tcolorbox}

\subsection{Evaluating reasoning steps using gpt-4o as a judge } 
\label{appendix:evaluating_resoning_steps_using_gpt4o_as_judge}

The \emph{evaluate\_steps} function is designed to rigorously assess the quality of reasoning steps generated by models against ground truth data using the GPT-4o-mini model. It takes the task question, ground truth reasoning, and model response as inputs and processes them within a structured conversation context. By leveraging a predefined system prompt and parameters like deterministic temperature (0.0) and a maximum token limit of 500, the function ensures consistent and reliable evaluations. The output provides clear feedback on alignment, logical flow, and coherence of reasoning steps, enabling precise analysis of model performance. This automated and standardized approach enhances objectivity, reproducibility, and detailed insight into multimodal reasoning capabilities.

\begin{tcolorbox}[colback=gray!5, colframe=black!45,
    fonttitle=\bfseries, title=Reasoning steps evaluation using gpt-4o as a judge
    ]
\small

\lstset{breaklines=true, basicstyle=\small\ttfamily} 

\begin{lstlisting}
def evaluate_steps(question , ground_truth, llm_response):
    messages = [
        {"role": "system", "content": system_prompt},
        {
            "role": "user",
            "content": [
                {"type": "text", "text": question + "\n" + f"Ground Truth : {ground_truth}" + "\n" + f"LLM Response : {llm_response}" },
            ],
        }
    ]
    response = client.chat.completions.create(
        model="gpt-4o-mini",  
        messages=messages,
        response_format=response_format,
        max_tokens=500,
        temperature = 0.0,
    )
    return response.choices[0].message.content
\end{lstlisting}

\end{tcolorbox}

\newpage

\subsection{Evaluating final answer accuracy } 
\label{appendix:evaluating_final_answer_accuracy}

To objectively assess how well the model’s final answer predictions align with the ground truth, we developed a comparison function that utilizes a secondary system prompt to evaluate response accuracy. This function analyzes the semantic similarity between the ground truth and the model's output, assigning a binary score: 1 for a match and 0 for a mismatch. By exclusively producing numeric scores, this approach ensures a precise and quantifiable evaluation of the model’s performance, effectively complementing the structured framework outlined earlier.

\begin{tcolorbox}[colback=gray!5, colframe=black!45,
    fonttitle=\bfseries, title=Secondary system prompt to evaluate the final answer accuracy
    ]
\small

\lstset{breaklines=true, basicstyle=\small\ttfamily} 

\begin{lstlisting}
system_prompt_2 = """
    You are a helpful Assistant. Provide helpful 
    response to the user's question.
"""

\end{lstlisting}

\end{tcolorbox}

\begin{tcolorbox}[colback=gray!5, colframe=black!45,
    fonttitle=\bfseries, title=Evaluate the Final Answer
    ]
\small

\lstset{breaklines=true, basicstyle=\small\ttfamily} 

\begin{lstlisting}
def compare_results(question, ground_truth, llm_response):
    messages = [
        {"role": "system", "content": system_prompt_2},
        {
            "role": "user",
            "content": [
                {
                "type": "text", 
                "text": f"""   
                    Evaluate the following answer based on 
                    Accuracy:
                    Question: {question}
                    Ground Truth: {ground_truth}
                    Model Prediction: {llm_response}
                    Match the meaning of the ground truth with the 
                    model prediction and if it matches give a 1. 
                    Otherwise 0.
                    Strictly return only the numeric score, 
                    without any additional commentary
                """
                },
            ],
        }
    ]

    response = client.chat.completions.create(
        model="gpt-4o-mini",  
        messages=messages,
        max_tokens=10,
        temperature=0.0
    )

    return response.choices[0].message.content

\end{lstlisting}

\end{tcolorbox}


\end{document}